\documentclass[runningheads]{llncs}

\usepackage[mobile]{eccv}

\usepackage{eccvabbrv}

\usepackage{graphicx}
\usepackage{booktabs}

\usepackage[accsupp]{axessibility}  

\usepackage{hyperref}

\usepackage{orcidlink}

\usepackage{multirow}
\usepackage{bbm}

\usepackage{amsmath,amsfonts,bm}

\def\eqref#1{equation~\ref{#1}}

\def\1{\bm{1}}

\DeclareMathAlphabet{\mathsfit}{\encodingdefault}{\sfdefault}{m}{sl}
\SetMathAlphabet{\mathsfit}{bold}{\encodingdefault}{\sfdefault}{bx}{n}

\DeclareMathOperator*{\argmax}{arg\,max}

\begin{document}

\title{Understanding Robustness of Visual State Space Models for Image Classification} 

\titlerunning{Understanding Robustness of Visual SSMs}

\author{Chengbin Du,
Yanxi Li,
Chang Xu}

\authorrunning{C. Du et al.}

\institute{
    School of Computer Science, University of Sydney\\
    \email{\{cbdu5632, yali0722\}@uni.sydney.edu.au, c.xu@sydney.edu.au}
}
\maketitle

\begin{abstract}
    Visual State Space Model (VMamba) has recently emerged as a promising architecture, exhibiting remarkable performance in various computer vision tasks. However, its robustness has not yet been thoroughly studied.
    In this paper, we delve into the robustness of this architecture through comprehensive investigations from multiple perspectives.
    %
    Firstly, we investigate its robustness to adversarial attacks, employing both whole-image and patch-specific adversarial attacks. Results demonstrate superior adversarial robustness compared to Transformer architectures while revealing scalability weaknesses.
    Secondly, the general robustness of VMamba is assessed against diverse scenarios, including natural adversarial examples, out-of-distribution data, and common corruptions. VMamba exhibits exceptional generalizability with out-of-distribution data but shows scalability weaknesses against natural adversarial examples and common corruptions.
    Additionally, we explore VMamba's gradients and back-propagation during white-box attacks, uncovering unique vulnerabilities and defensive capabilities of its novel components.
    Lastly, the sensitivity of VMamba to image structure variations is examined, highlighting vulnerabilities associated with the distribution of disturbance areas and spatial information, with increased susceptibility closer to the image center.
    Through these comprehensive studies, we contribute to a deeper understanding of VMamba's robustness, providing valuable insights for refining and advancing the capabilities of deep neural networks in computer vision applications.
\end{abstract}

\section{Introduction}
\label{sec:introduction}

    Deep neural networks represent a cornerstone of contemporary research \cite{han2022ghostnets, wang2018learning}, but their robustness in the face of adversarial attacks \cite{goodfellow2014explaining, madry2017towards, fu2022patch} and other perturbations \cite{hendrycks2021nae, hendrycks2021many, hendrycks2019robustness} remains a critical concern. Researchers are increasingly focused on developing models that not only excel in specific tasks but also demonstrate resilience against adversarial attacks while maintaining strong generalization capabilities across diverse scenarios.
    Recently, a novel and promising addition to the landscape of neural network architectures for visual representation learning has emerged, known as the Visual State Space Model \cite{liu2024vmamba, zhu2024vision, huang2024localmamba}.
    This architecture has earned attention for its exceptional performance across a spectrum of computer vision tasks.
    Despite the considerable successes in various applications, an aspect that has not yet been thoroughly studied is the robustness of VMamba.

    This paper addresses the existing gap in the understanding of VMamba's robustness by undertaking a comprehensive investigation.
    To comprehensively evaluate the robustness of VMamba, our analysis takes a multi-faceted approach, thoroughly exploring the robustness of VMamba from various perspectives.
    


    Firstly, we analyze the robustness of VMamba against adversarial attacks. 
    We employ two types of adversarial attacks. The first type targets the entire image, which includes Fast Gradient Sign Method (FGSM) \cite{goodfellow2014explaining} and Projected Gradient Descent (PGD) \cite{madry2017towards}. The second type focuses on attacking only several patches in an image, which includes Patch-Fool \cite{fu2022patch}. This analysis reveals:
    \begin{itemize}
        \item[$\bullet$] VMamba has better adversarial robustness than Transformer architectures \cite{steiner2021train, pmlr-v139-touvron21a, liu2021swin}.
        \item[$\bullet$] The scalability of VMamba proves relatively weak against adversarial attacks.
    \end{itemize}
    Secondly, we assess the general robustness of VMamba, evaluating its performance against natural adversarial examples in ImageNet-A \cite{hendrycks2021nae}, out-of-distribution data in ImageNet-R \cite{hendrycks2021many}, and common corruptions in ImageNet-C \cite{hendrycks2019robustness}. Understanding the model's behavior in these diverse scenarios is crucial for establishing its reliability in real-world applications. This analysis reveals:
    \begin{itemize}
        \item[$\bullet$] VMamba exhibits superior generalizability when faced with out-of-distribution data.
        \item[$\bullet$] The scalability of VMamba proves relatively weak against natural adversarial examples and common corruptions.
    \end{itemize}
    Furthermore, we perform experiments to inspect the gradients of VMamba and the back-propagation process within VMamba when subjected to white-box attacks. This exploration aims to investigate how the novel components in VMamba behave under adversarial attacks. These novel components have not been previously observed in architectures designed for vision tasks. This analysis reveals:
    \begin{itemize}
        \item[$\bullet$] Gradients of the parameter $A$ is hard to be estimated by the attack algorithm.
        \item[$\bullet$] Gradients of the parameters $B$ and $C$ primarily contribute to the VMamba vulnerability, which exhibits increased vulnerability correlated to the increasing of the model size.
        \item[$\bullet$] Parameter \(\Delta\) demonstrates defensive capabilities against white-box attacks, with its effectiveness growing with model size.
        \item[$\bullet$]The trade-off between the parameters $B$, $C$, and \(\Delta\) leads to the robustness of the VMamba model not increasing proportionally with the model size.
    \end{itemize}
    Finally, we turn our attention to exploring the sensitivity of VMamba to the structures of images.
    This investigation encompasses a comparative analysis involving the random removal of image patches and pixels.
    Additionally, we introduce permutations in the order of image patches to assess their impact.
    We also employ the Patch-Fool to systematically attack specific positions within the image patches.
    This multifaceted examination aims to provide a nuanced understanding of how VMamba responds to variations in image structure, offering valuable insights into its robustness and potential vulnerabilities.
    This analysis reveals:
    \begin{itemize}
        \item[$\bullet$] VMamba model is sensitive to the continuity in the scanning trajectory.
        \item[$\bullet$] Vmamba model has a broader receptive field than Swin model.
        \item[$\bullet$] With an equal amount of perturbation, the more patches that are affected, the more vulnerable the VMamba model becomes.
        \item[$\bullet$] VMamba is highly sensitive to the spatial information of images.
        \item[$\bullet$] The closer the perturbation is to the center of the image, the more vulnerable VMamba will be.
    \end{itemize}

    Our research offers a comprehensive understanding of the robustness of VMamba from various perspectives.
    Through empirical analysis, we delve into the factors that may influence VMamba's robustness, shedding light on critical aspects that warrant attention for further refinement.
    By systematically examining its performance under different conditions and stressors, our study contributes valuable insights that can inform future developments in enhancing the VMamba architecture (as discussed in Section \ref{sec:insights}).
    The findings provides a roadmap for researchers to iteratively refine and optimize VMamba, ultimately advancing its robustness to achieve superior performance compared to its current state.

\section{Preliminaries}

    \subsection{Vision Transformers}
    
        The \textbf{Transformer} \cite{vaswani2017attention} is a model architecture that relies solely on attention mechanisms, initially designed for Natural Language Processing (NLP) tasks.
        Following the successes of Transformers in NLP tasks, Vision Transformer (\textbf{ViT}) \cite{dosovitskiy2020image, su2022vitas} explores the application of a standard Transformer directly to images with minimal modifications. The approach involves dividing an image into patches, treating them akin to tokens (words) in an NLP application, and presenting the sequence of linear embeddings of these patches as input to the Transformer.
        
        The \textbf{Swin} Transformer \cite{liu2021swin} introduces a novel hierarchical Transformer architecture with a distinctive emphasis on Shifted windows for representation computation. The proposed shifted windowing scheme enhances efficiency by confining self-attention computation to non-overlapping local windows, concurrently facilitating cross-window connections. This hierarchical design offers flexibility in modeling at multiple scales and maintains linear computational complexity concerning image size. The Swin Transformer's notable attributes, including its efficient windowing strategy, render it versatile and applicable across various vision tasks.

    \subsection{State Space Models}

        A novel category of sequence models for deep learning is emerging, which is known as Structured State Space Sequence (S4) models.
        Using an implicit latent state $h(t) \in \mathbb{R}^{N}$, S4 models can map a 1-dimensional function or sequence $x(t) \in \mathbb{R}^{L} \mapsto y(t) \in \mathbb{R}^{L}$:
        \begin{align}\label{eq:ode}
            \begin{split}
                h'(t) = \bm{A}h(t) + \bm{B}x(t), \quad\quad y(t) = \bm{C}h(t),
            \end{split}
        \end{align}
        where $\bm{A}\in\mathbb{R}^{N\times N}$, $\bm{B}\in\mathbb{R}^{N\times 1}$, and $\bm{C}\in\mathbb{R}^{N\times 1}$ are continuous parameters. 

        In practice, the continuous parameters in Eq. \ref{eq:ode} need to be first discretized.
        This can be achieved using a zero-order hold (ZOH):
        \begin{equation}
            \overline{\bm{A}} = \exp(\Delta \bm{A})
            \quad\quad
            \overline{\bm{B}} = (\bm{\Delta} \bm{A})^{-1}(\exp(\Delta \bm{A}) - \bm{I}) \cdot \Delta \bm{B},
            \label{eq:preliminaries:zoh}
        \end{equation}
        where $\bm{\overline{A}}$, $\bm{\overline{B}}$ are the discrete counterparts of the continuous parameters $\bm{A}$ and $\bm{B}$, and $\bm{\Delta} \in \mathbb{R}{>0}$ is a specified sampling timescale for the discretization.
        The discretization leads to a discretized form of the model as follows:
        \begin{align} \label{eq:discretization}
            \begin{split}
                h_t = \bm{\overline{A}}h_{t-1} + \bm{\overline{B}}x_t, \quad\quad y_t = \bm{C}h_t.
            \end{split}
        \end{align}
        A remaining issue is that the iterative process in Eq. \ref{eq:discretization} is not computationally efficient. To enhance efficiency, it can be sped up through parallel computation. With a global convolution operation (denoted by $\circledast$), we obtain:
        \begin{align}
            \begin{split}\label{eq:Cumulative}
                \bm{y} &= \bm{x} \circledast \bm{\overline{K}} \\
                \text{with} \quad \bm{\overline{K}} &= (\bm{C}\bm{\overline{B}},\bm{C}\overline{\bm{A}\bm{B}}, ..., \bm{C}\bm{\overline{A}}^{L-1}\bm{\overline{B}}),
            \end{split}
        \end{align}
        where $\bm{\overline{K}} \in \mathbb{R}^L$ is a kernel used in the S4 model. 
        This method uses convolution to generate outputs across the sequence at the same time, improving computational efficiency and scalability.

    \subsection{Selective State Space Models}

        Traditional State Space Models (S4) are known for their linear time complexity but face limitations in capturing sequence context due to fixed parameterization. The Mamba models \cite{gu2023mamba}, overcome these limitations by implementing a dynamic and selective approach for managing interactions between sequential states. Unlike standard SSMs that rely on constant transition parameters (\(\bm{\overline{A}}, \bm{\overline{B}}\)), Mamba utilizes parameters that depend on the input, enabling more complex, sequence-aware parameterization. This approach involves directly deriving parameters \(\bm{B}, \bm{C},\) and \(\bm{\Delta}\) from the input sequence \(\bm{x}\), which allows for a richer representation of sequence context.

        By adopting selective SSMs, Mamba models not only maintain linear scalability with respect to sequence length but also demonstrate strong performance in language modeling tasks. This innovation has paved the way for their application in vision tasks as well, inspiring the development of new models that integrate Mamba. For example, Vim \cite{zhu2024vision} combines Mamba with a ViT-like structure by including bi-directional Mamba blocks instead of the usual Transformer blocks. Similarly, VMamba \cite{liu2024vmamba} presents an innovative 2D selective scanning method for processing images in both horizontal and vertical directions and builds a hierarchical model that is reminiscent of the Swin Transformer \cite{liu2021swin}. 

\section{Adversarial Robustness}

    \subsection{White-box Attacks}
    \label{sec:white_box}
    
        We employ two types of adversarial attacks. The first type targets the entire image, which includes Fast Gradient Sign Method (FGSM) \cite{goodfellow2014explaining} and Projected Gradient Descent (PGD) \cite{madry2017towards}.
        We employ an $l_{\infty}$-norm with a perturbation magnitude of $\varepsilon=1/255$ for both FGSM and PGD. FGSM operates in a single step, whereas PGD represents a multi-step variant of FGSM. Specifically, we iterate PGD for 5 steps, utilizing a step size of $0.5/255$.
        The second type focuses on attacking only several patches in an image, which includes Patch-Fool \cite{fu2022patch}.
        In Patch-Fool experiments, the weight coefficient $\alpha$ is fixed at 0.002. The initial step size $\eta$ is set to 0.2 and undergoes a 0.95 decay every 10 iterations, with a total of 250 iterations.
        The targeted patch for the attack is chosen randomly from all the 196 patches. Our experiments include different numbers of targeted patches, ranging from 1 to 4. 
        The learning of adversarial noises is facilitated through the Adam optimizer. 
        For the efficiency of the Patch-Fool experiment, we randomly selected a subset of 2,500 images from the ImageNet validation dataset.

        \subsubsection{Attack the entire image.}
        The results under entire-image attacks are reported in Table \ref{tab:exp:main-results}.
        We majorly compare VMamba to the leading Transformer architecture, Swin, which exhibits similar performance to VMamba on clean images and demonstrates the best adversarial robustness among the three Transformer architectures under FGSM and PGD attacks.
        Based on the results, we can see:
        \begin{itemize}
            \item[$\bullet$] \textbf{VMamba has better adversarial robustness than Transformer architectures.}
            \item[$\bullet$] \textbf{The scalability of VMamba proves relatively weak against adversarial attacks.}
        \end{itemize}
        Comparing to Swin-B, VMamba-B and B* achieve 4.1\% and 4.3\% higher accuracy under FGSM, and 12.2\% and 13.1\% higher accuracy under PGD, respectively.

        \begin{table}[!tbp]
        \centering
        \caption{Evaluation of SOTA methods on ImageNet-1K. 
        The top-1 accuracy is used to assess performance on clean ImageNet-1K and under adversarial attacks (FGSM and PGD).
        All models utilize input dimensions of $224\times224$.}
        \begin{tabular}{c|l|c|cc}
            \toprule
                \textbf{~Categories~} & 
                \textbf{Models} & 
                \textbf{~Clean~} & 
                \textbf{~FGSM~} &
                \textbf{~PGD~} \\
            \midrule
                \multirow{7}{*}{Transformer}
                & ViT-S/16 + AugReg \cite{steiner2021train} & 74.7 & 27.2 & ~9.3 \\ 
                & ViT-B/16 + AugReg \cite{steiner2021train} & 76.8 & 42.0 & 23.1 \\ 
                & DeiT-Ti \cite{pmlr-v139-touvron21a}        & 72.2 & 22.3 & ~6.1 \\
                & DeiT-S \cite{pmlr-v139-touvron21a}        & 79.8 & 40.5 & 16.4 \\ 
                & DeiT-B \cite{pmlr-v139-touvron21a}        & 81.8 & 46.3 & 21.6 \\ 
                & Swin-T \cite{liu2021swin}                 & 81.2 & 33.8 & 7.30 \\ 
                & Swin-S \cite{liu2021swin}                 & 83.2 & 45.9 & 18.3 \\ 
                & Swin-B \cite{liu2021swin}                 & 83.5 & 49.8 & 21.6 \\ 
            \midrule
                \multirow{4}{*}{\begin{tabular}{@{}c@{}}SSM\end{tabular}}
                & VMamba-T  \cite{liu2024vmamba}            & 82.2 & 48.6 & 28.3 \\ 
                & VMamba-S  \cite{liu2024vmamba}            & 83.5 & 53.4 & 34.5 \\ 
                & VMamba-B  \cite{liu2024vmamba}            & 83.2 & 53.9 & 33.8 \\ 
                & VMamba-B* \cite{liu2024vmamba}            & 83.7 & 54.1 & 34.7 \\ 
            \bottomrule
        \end{tabular}
        \label{tab:exp:main-results}
        \end{table}

        The experiment analysis highlights notable trends in the adversarial robustness and scalability of VMamba models compared to Swin counterparts.
        Across various attack scenarios, VMamba consistently outperforms Swin models, showcasing superior adversarial robustness. The discrepancies are particularly pronounced in the smaller models, with VMamba-T and VMamba-S exhibiting 14.8\% and 21.0\%, and 7.5\% and 16.2\% higher accuracy than Swin-T and Swin-S under FGSM and PGD attacks, respectively.
        Moreover, the larger VMamba-B and VMamba-B* models demonstrate robustness advantages of 4.1\% and 4.3\% under FGSM, and a substantial 12.2\% and 13.1\% under PGD attacks when compared to Swin-B. However, the analysis also reveals a weakness in VMamba's scalability, as the performance gap narrows with increasing model size. This trend is evident both in adversarial scenarios and clean images, indicating that VMamba's advantages diminish as the model complexity grows. The extension to this observation underscores that VMamba's scalability challenges persist even on clean images, though the effect is more pronounced under adversarial attacks.

        \begin{table}[!tbp]
        \centering
        \caption{Robust Accuracy under Patch-fool attack, "P1" to "P4" represent the robust accuracy when a certain number of patches are under attack. P1-P4 signifies the difference in robust accuracy between P1 and P4.}
        \label{tab:Number_patches}
        \begin{tabular}{c|l|c|cccc|c}
            \toprule
                \textbf{~Categories~} &
                \textbf{Model} &
                \textbf{~Clean~} &
                \textbf{~P1~} &
                \textbf{~P2~} &
                \textbf{~P3~} &
                \textbf{~P4~} &
                \textbf{~P1$-$P4~} \\
            \midrule
                \multirow{7}{*}{Transformer}
                & ViT-S/16 + AugReg \cite{steiner2021train} & 75.9 & ~0.1 & ~0.0 & ~0.0 & ~0.0 & ~0.1 \\
                & ViT-B/16 + AugReg \cite{steiner2021train} & 78.3 & 12.2 & ~0.9 & ~0.0 & ~0.0 & 12.2 \\
                & DeiT-T \cite{pmlr-v139-touvron21a}        & 72.6 & ~4.9 & ~0.0 & ~0.0 & ~0.0 & ~4.9 \\
                & DeiT-S \cite{pmlr-v139-touvron21a}        & 81.4 & ~6.6 & ~0.2 & ~0.0 & ~0.0 & ~6.6 \\
                & DeiT-B \cite{pmlr-v139-touvron21a}        & 81.9 & 25.4 & ~1.7 & ~0.0 & ~0.0 & 25.4 \\
                & Swin-T \cite{liu2021swin}                 & 81.5 & 40.3 & 16.6 & ~5.7 & ~2.2 & 38.1 \\
                & Swin-S \cite{liu2021swin}                 & 84.0 & 52.6 & 22.9 & 10.7 & ~6.0 & 46.6 \\
                & Swin-B \cite{liu2021swin}                 & 84.5 & 43.7 & 15.7 & ~5.1 & ~2.2 & 41.6 \\
            \midrule
                \multirow{4}{*}{\begin{tabular}{@{}c@{}}SSM\end{tabular}}
                & VMamba-T  \cite{liu2024vmamba} & 83.1 & 50.7 & 24.8 & 17.7 & 11.5 & 39.2 \\
                & VMamba-S  \cite{liu2024vmamba} & 84.7 & 59.9 & 35.3 & 21.5 & 13.7 & 46.2 \\
                & VMamba-B  \cite{liu2024vmamba} & 83.7 & 59.7 & 30.8 & 16.2 & 10.2 & 49.6 \\
                & VMamba-B* \cite{liu2024vmamba} & 84.9 & 58.1 & 26.7 & 14.5 & ~7.2 & 50.9 \\
            \bottomrule
        \end{tabular}
        \end{table}

        \subsubsection{The patch-wise attack.}
        We randomly select 2500 images from the validation set of ImageNet for evaluating patch-wise robustness, following \cite{fu2022patch}. The results under patch-wise attacks are reported in Table \ref{tab:Number_patches}.
        The "Clean" column indicates the accuracy with clean images, while columns "P1" to "P4" represent the robust accuracy when a certain number of patches are under attack. P1-P4 signifies the difference in robust accuracy between P1 and P4. 
        
        Compared to the Swin models, the VMamba model maintains a higher accuracy within the P1 to P4 range, suggesting that the VMamba model exhibits superior robustness against patch-wise white-box adversarial attacks. Although the robust accuracy of all models shows a downward trend as the number of attacked patches increases from P1 to P4, it's noteworthy that the VMamba model exhibits a more pronounced P1$-$P4 difference compared to the Swin model except the VMamba-S. For instance, VMamba-T, VMamba-B, and VMamba-B* exhibited 1.1\%, 8.0\%, and 9.3\% higher accuracy than Swin-T and Swin-B. This phenomenon illustrates that compared with the Swin model, the robustness of VMamba is more dependent on the number of disturbed patches and probably related to model size. We conduct a more detailed analysis of this perspective in section \ref{sec:Information-loss}.

    \subsection{Black-box Attacks and Transferability of Noises}

        In this section, we comprehensively assessed the transferability of adversarial samples between the VMamba and transformer models. In Table \ref{tab:Blackbox}, the first column "A vs. B" indicates the two models used for assessment. The second and third columns respectively show the robust accuracy of the A and B models on clean images. The fourth column, "Left to Right", represents the robust accuracy after transferring the adversarial samples generated using model A's gradients to model B, and vice versa. 
        
        From Table \ref{tab:Blackbox}, we can see that when using adversarial samples generated by the Swin model to transfer attack VMamba, all model sizes of VMamba exhibited the lowest robust accuracy rates, with VMamba-T, VMamba-S, VMamba-B, and VMamba-B* being 73.9\%, 76.1\%, 74.0\%, and 75.5\% respectively. This phenomenon indicates that compared to other transformer models, VMamba and Swin are more similar in their behavior of extracting image features due to their identical hierarchical model structures. Therefore, in order to fairly compare the robustness differences between SSM and Transformer models, we prefer to compare VMamba with Swin in the following sections. In addition, the discussion between the Vim and Transformer models will be provided in supplementary material.

        \begin{table}[!tbp]
            \centering
            \caption{Transferability of adversarial samples between the SSM and transformer models. The first column "A vs. B" indicates the two models used for assessment. The second and third columns respectively show the robust accuracy of the A and B models on clean images. The fourth column, "Left to Right", represents the robust accuracy after transferring the adversarial samples generated using model A's gradients to model B, and vice versa. All adversarial samples are generated by using PGD. Vim\textsuperscript{\textdagger} is the version adapted for Long Sequence Fine-tuning. Specifically, they maintained the original patch size but adjusted the patch extraction stride to 8.}
            \label{tab:Blackbox}
            \begin{tabular}{l|cc|c|c}
                \toprule
                    \textbf{} & \textbf{SSM} &
                    \textbf{Transformer} & \textbf{Left $\rightarrow$ Right} & \textbf{Right $\rightarrow$ Left} \\
                \midrule
                    VMamba-T vs. DeiT-Ti & 82.2 & 72.2 & 69.0 & 79.4   \\
                    VMamba-T vs. Swin-T  & 82.2 & 81.2 & 68.4 & \textbf{73.9}  \\
                \midrule
                    VMamba-S vs. DeiT-S & 83.5 & 79.8 & 76.1 & 79.6   \\
                    VMamba-S vs. ViT-S & 83.5 & 74.7 & 71.8 & 80.8   \\
                    VMamba-S vs. Swin-S & 83.5 & 83.2 & 69.7 & \textbf{76.1}   \\

                \midrule    
                    VMamba-B vs. DeiT-B & 83.2 & 81.8 & 78.4 & 78.9   \\
                    VMamba-B vs. ViT-B & 83.2 & 76.8 & 74.7 & 80.4   \\
                    VMamba-B vs. Swin-B & 83.2 & 83.5 & 80.0 & \textbf{74.9}   \\
                    VMamba-B* vs. DeiT-B & 83.7 & 81.8 & 77.7 & 79.4   \\
                    VMamba-B* vs. ViT-B & 83.7 & 76.8 & 74.2 & 80.8  \\
                    VMamba-B* vs. Swin-B & 83.7 & 83.5 & 68.8 & \textbf{75.5}   \\

                \bottomrule
            \end{tabular}
        \end{table}
        
    \subsection{General Robustness}
        The ImageNet-A dataset \cite{hendrycks2021nae} comprises natural adversarial examples that challenge models by placing ImageNet objects in unconventional contexts or orientations. This assesses the model's adaptability to unexpected scenarios.
        In contrast, the ImageNet-R dataset \cite{hendrycks2021many} introduces out-of-distribution data, presenting abstract or rendered versions of objects to test the model's ability to generalize beyond its trained data distribution.
        Lastly, the ImageNet-C dataset \cite{hendrycks2019robustness} introduces common corruptions, incorporating 19 distortions across 5 categories, such as motion blur, Gaussian noise, fog, and JPEG compression. This dataset emulates real-world distortions, providing insights into a model's resilience to diverse environmental challenges.
        The results are reported in Table \ref{tab:exp:main-results-variants}.

        \begin{table}[!tbp]
        \centering
        \caption{Evaluation of SOTA methods on ImageNet variants (A, R and C). 
        The top-1 accuracy is used to assess performance on ImageNet-A, and -R. In the case of ImageNet-C, the focus is on the mean Corruption Error (mCE), where lower values indicate better performance (marked by ${\downarrow}$). 
        All models utilize input dimensions of $224\times224$.}
        \begin{tabular}{c|l|ccc}
            \toprule
                \textbf{~Categories~} &
                \textbf{Models} &
                \textbf{~~A~~} & \textbf{~~R~~} & \textbf{~C} ${(\downarrow)}$ \\
            \midrule
                \multirow{7}{*}{Transformer}
                & ViT-S/16 + AugReg \cite{steiner2021train} & ~9.0  & 31.9 & 53.4 \\
                & ViT-B/16 + AugReg \cite{steiner2021train} & 11.7 & 36.9 & 47.8 \\
                & DeiT-T    \cite{pmlr-v139-touvron21a}     & 7.6 & 32.7 & 53.6 \\
                & DeiT-S    \cite{pmlr-v139-touvron21a}     & 19.5 & 41.9 & 41.2 \\
                & DeiT-B    \cite{pmlr-v139-touvron21a}     & 27.8 & 44.6 & 36.7 \\
                & Swin-T    \cite{liu2021swin}              & 21.2 & 41.2 & 45.9 \\
                & Swin-S    \cite{liu2021swin}              & 32.6 & 44.8 & 41.0 \\
                & Swin-B    \cite{liu2021swin}              & 36.0 & 46.4 & 40.2 \\
            \midrule
                \multirow{4}{*}{\begin{tabular}{@{}c@{}}SSM\end{tabular}}
                & VMamba-T     \cite{liu2024vmamba}         & 27.0 & 45.5 & 40.0 \\
                & VMamba-S     \cite{liu2024vmamba}         & 32.7 & 50.5 & 35.5 \\
                & VMamba-B     \cite{liu2024vmamba}         & 35.4 & 49.4 & 35.7 \\
                & VMamba-B*    \cite{liu2024vmamba}         & 36.8 & 51.4 & 35.0 \\
            \bottomrule
        \end{tabular}
        \label{tab:exp:main-results-variants}
        \end{table}

        Analyzing the results on ImageNet-R reveals that:
        \begin{itemize}
            \item[$\bullet$] \textbf{VMamba exhibits superior generalizability when faced with out-of-distribution data (ImageNet-R).}
        \end{itemize}
            Notably, VMamba-T, S, B, and B* models consistently exhibits superior performance when compared to state-of-the-art Transformer-based models on ImageNet-R.
            VMamba-T surpasses Swin-T by 4.3\%, VMamba-S outperforms Swin-S by 5.7\%, VMamba-B exceeds Swin-B by 3.0\%, and VMamba-B* is 5.0\% higher than Swin-B.

        However, results on ImageNet-A and C reveals a weakness of VMamba:
        \begin{itemize}
            \item[$\bullet$] \textbf{The scalability of VMamba proves relatively weak against natural adversarial examples (ImageNet-A) and common corruptions (ImageNet-C).}
        \end{itemize}
        Despite the notable performance of the smallest variant, VMamba-T, its larger counterparts, namely VMamba-S, B, and B*, fail to maintain this superiority.
        While VMamba-T showcases a substantial superiority of 5.8\% and 5.9\% on ImageNet-A and C, respectively, the larger VMamba-S and B models do not exhibit the same level of consistency. VMamba-S outperforms Swin-S on ImageNet-C by 5.6\% but only marginally improves on ImageNet-A (0.1\%).
        VMamba-B and VMamba-B* show a less efficient scaling, being merely around 1\% better or even worse than the best Transformer-based B-sized models on ImageNet-A and C. This underscores the nuanced nature of VMamba's scalability, indicating that its performance may not uniformly improve with model size across different ImageNet variants.

    \section{Why is VMamba Robust to White-box Attacks?}
    \label{sec:ABCD}
        To analyze the robustness of the VMamba model under white-box adversarial attacks, we examine the effect of individual parameters on the robustness of models. This will be achieved by measuring the robust accuracy of models using PGD after individually deactivating the gradients of parameters $A$, $B$, $C$, and \(\Delta\). Table \ref{tab:ABCD} presents the robust accuracy of four variations of the VMamba model (T, S, B, and B*) on clean images and under PGD attacks, with appended data columns quantifying the robust accuracy upon the gradient inactivation of each respective parameter.

        \begin{table}[!tbp]
            \centering
            \caption{Robust accuracy without gradients of the parameters \(A\), \(B\), \(C\), or \(\Delta\)}
            \label{tab:ABCD}
            \begin{tabular}{l|cc|cccc}
                \toprule
                    \multirow{2}{*}{\textbf{Model}} &
                    \multirow{2}{*}{\textbf{~Clean~}} &
                    \multirow{2}{*}{\textbf{~PGD~}} &
                    \multicolumn{4}{c}{\textbf{w/o gradients of}} \\
                \cline{4-7}
                    &&&
                    ~~\(A\)~~ &
                    ~~\(B\)~~ &
                    ~~\(C\)~~ &
                    ~~\(\Delta\)~~\rule{0pt}{2.5ex}\\
                \midrule
                    VMamba-T  & 82.2 & 28.3 & 28.3 & 32.7 & 29.6  & 31.9  \\
                    VMamba-S  & 83.5 & 34.5 & 34.5 & 45.6 & 34.2  & 27.8  \\
                    VMamba-B  & 83.2 & 33.8 & 33.8 & 41.8 & 42.9  & 23.2  \\
                    VMamba-B* & 83.7 & 34.7 & 34.7 & 42.6 &  42.4  & 25.5  \\
                \bottomrule
            \end{tabular}
        \end{table}

        \begin{itemize}
            \item[$\bullet$] \textbf{Gradients of the parameter $A$ is hard to be estimated by the attack algorithm.}
        \end{itemize}

        Table \ref{tab:ABCD} exhibits that when the gradient of parameter A is deactivated, the robust accuracy of all VMamba models under PGD attack remains unchanged. This indicates that the PGD attack is unable to capture the dynamics controlled by parameter $A$ in the model.
        Upon closer examination of Equation \ref{eq:Cumulative}, it can be observed that within the scanning trajectory, the intermediate latent state $h(t)$ is updated at each step by incorporating the previous state through multiplication by the same parameter $A$. Empirically, this can cause the gradients of $A$ to become hard to be estimated by the PGD, thereby potentially stabilizing the model against such gradient-based attacks.

        \begin{itemize}
            \item[$\bullet$] \textbf{Gradients of the parameters $B$ and $C$ primarily contribute to the VMamba vulnerability, which exhibits increased vulnerability correlated to the increasing of the model size.}
        \end{itemize}
        
        By deactivating gradients of the parameter $B$, all VMmamba models consistently show a trend of increased robustness. For example, compared to standard PGD attack conditions, VMamba-T, VMamba-S, VMamba-B, and VMamba-B* improved by 4.4\%, 11.1\%, 8\%, and 7.9\%, respectively. This means that parameter $B$ plays a crucial role in the gradient estimation of PGD, where its gradient information provides exploitable weaknesses for PGD. Notably, from the overall trend, the vulnerability of parameter $B$ is positively correlated with the model size, that is, both the Small and Base versions exhibit higher vulnerability than the Tiny version. 
        For parameter $C$, deactivating its gradients only significantly improves the robustness of models on VMamba-B and B*, with robust accuracy increasing by 9.1\% and 7.7\%, respectively. Conversely, VMamba-T exhibited a marginal improvement of 1.3\%, and VMamba-S demonstrated a slight reduction in robustness, quantified as a 0.3\% decrease.
        Both phenomenons suggest that as the model size increases, the model becomes more complex and nonlinear, making parameters B and C more sensitive to perturbations in the input.
        
        \begin{itemize}
            \item[$\bullet$] \textbf{The parameter \(\Delta\) demonstrates defensive capabilities against white-box attacks, with its effectiveness growing with model size.}
        \end{itemize}
        
        In the VMamba model, the \(\Delta\) parameter transforms continuous-time system parameters into their discrete counterparts, enhancing the model's ability to efficiently process sequences over specified sampling timescales. By deactivating the gradient of parameter \(\Delta\), the robust accuracy of the VMamba-T model increased from 28.3\% to 31.9\%. In contrast, there was a significant decrease in the robust accuracy for the VMamba-S, VMamba-B, and VMamba-B* models, which dropped by 6.7\%, 10.6\%, and 9.2\%, respectively. This phenomenon suggests that the \(\Delta\) parameter can defend against white-box attacks only when the model size reaches at least the level of Small, and this ability tends to steadily increase as the model size grows. On the other hand, the discretization facilitated by \(\Delta\) enables the model to more effectively capture and integrate temporal dependencies over specified intervals, thereby incorporating information across time to enhance robustness against temporal variations and noise.

        \begin{figure}[!tbp]
            \centering
            \begin{subfigure}[b]{0.33\textwidth}
                \centering
                \includegraphics[width=0.9\textwidth]{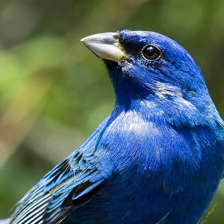}
                \caption{Origin Image}
                \label{fig:origin_image}
            \end{subfigure}%
            \hfill
            \begin{subfigure}[b]{0.33\textwidth}
                \centering
                \includegraphics[width=0.9\textwidth]{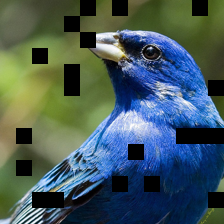}
                \caption{Patch-wise Drop}
                \label{fig:patch_case}
            \end{subfigure}%
            \hfill
            \begin{subfigure}[b]{0.33\textwidth}
                \centering
                \includegraphics[width=0.9\textwidth]{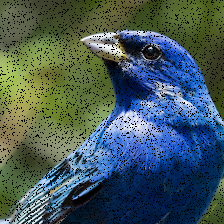}
                \caption{Pixel-wise Drop}
                \label{fig:pixel_case}
            \end{subfigure}%
            \caption{An example image with its Patch-wise drop and Pixel-wise drop version.}
            \label{fig:drop_case}
        \end{figure}

        \begin{itemize}
            \item[$\bullet$] \textbf{The trade-off between the parameters \( B \), \( C \), and delta leads to the robustness of the VMamba model not increasing proportionally with the model size.}
        \end{itemize}

        In summary, the white-box attack algorithms aim to learn optimal noises from the gradients of the model. They cannot effectively learn meaningful knowledge from the gradients of $A$ and $\Delta$. whereas the gradients of parameters \( B \) and \( C \) are identified as pivotal in the existing vulnerabilities. In contrast, the gradient of the parameter \( \Delta \) exhibits a protective mechanism, establishing a delicate balance that will vary with the size of the model. Thus, unlike the Swin model, the robustness of the VMamba model does not increase proportionally with model size.

    \section{Sensitivity to Information Loss}
    \label{sec:Information-loss}

        \begin{figure}[!tbp]
            \centering
            \begin{subfigure}[b]{0.245\textwidth}
                \centering
                \includegraphics[width=0.9\textwidth]{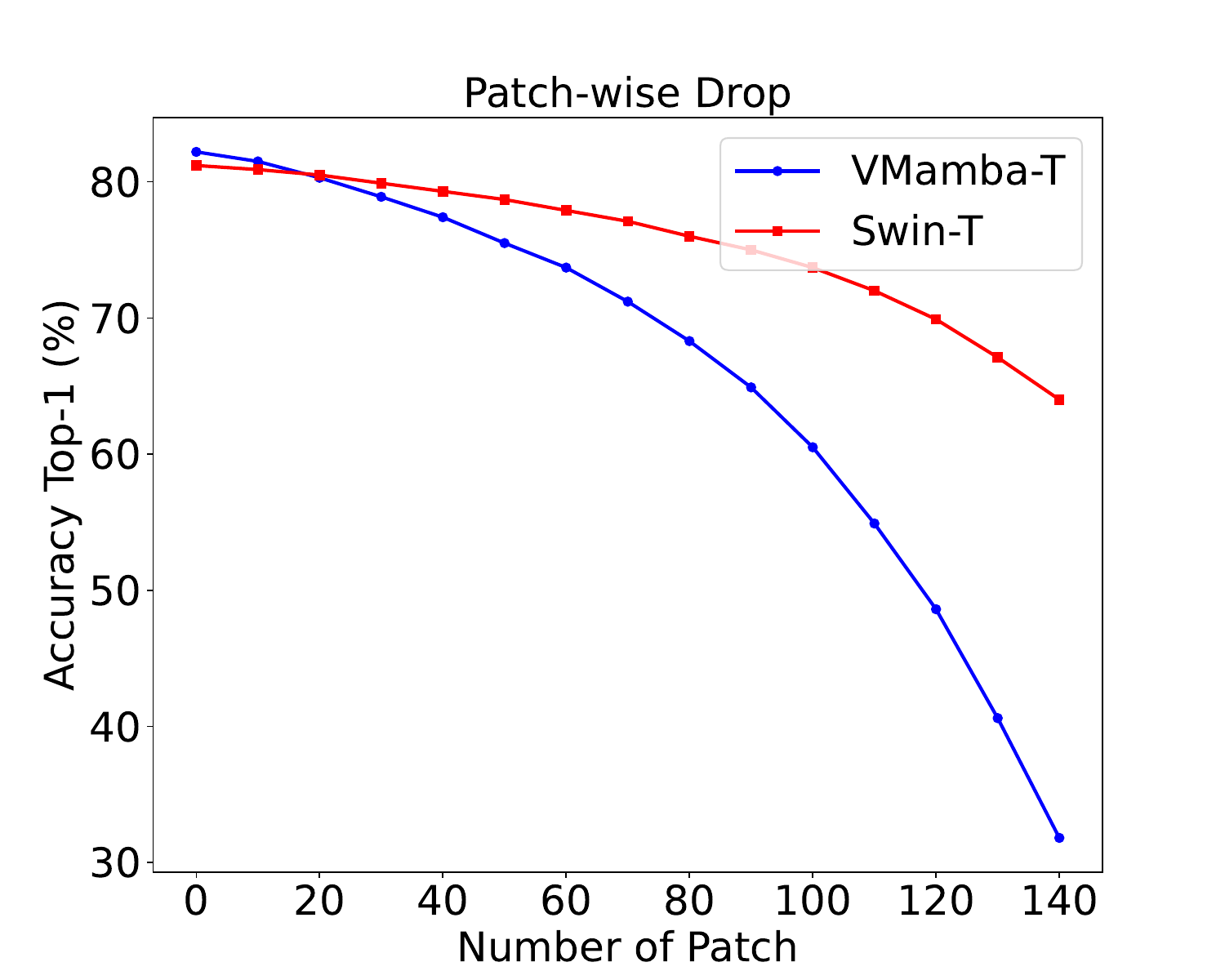}
                \caption{Tiny}
                \label{fig:RandomDrop_T}
            \end{subfigure}
            \hfill
            \begin{subfigure}[b]{0.245\textwidth}
                \centering
                \includegraphics[width=0.9\textwidth]{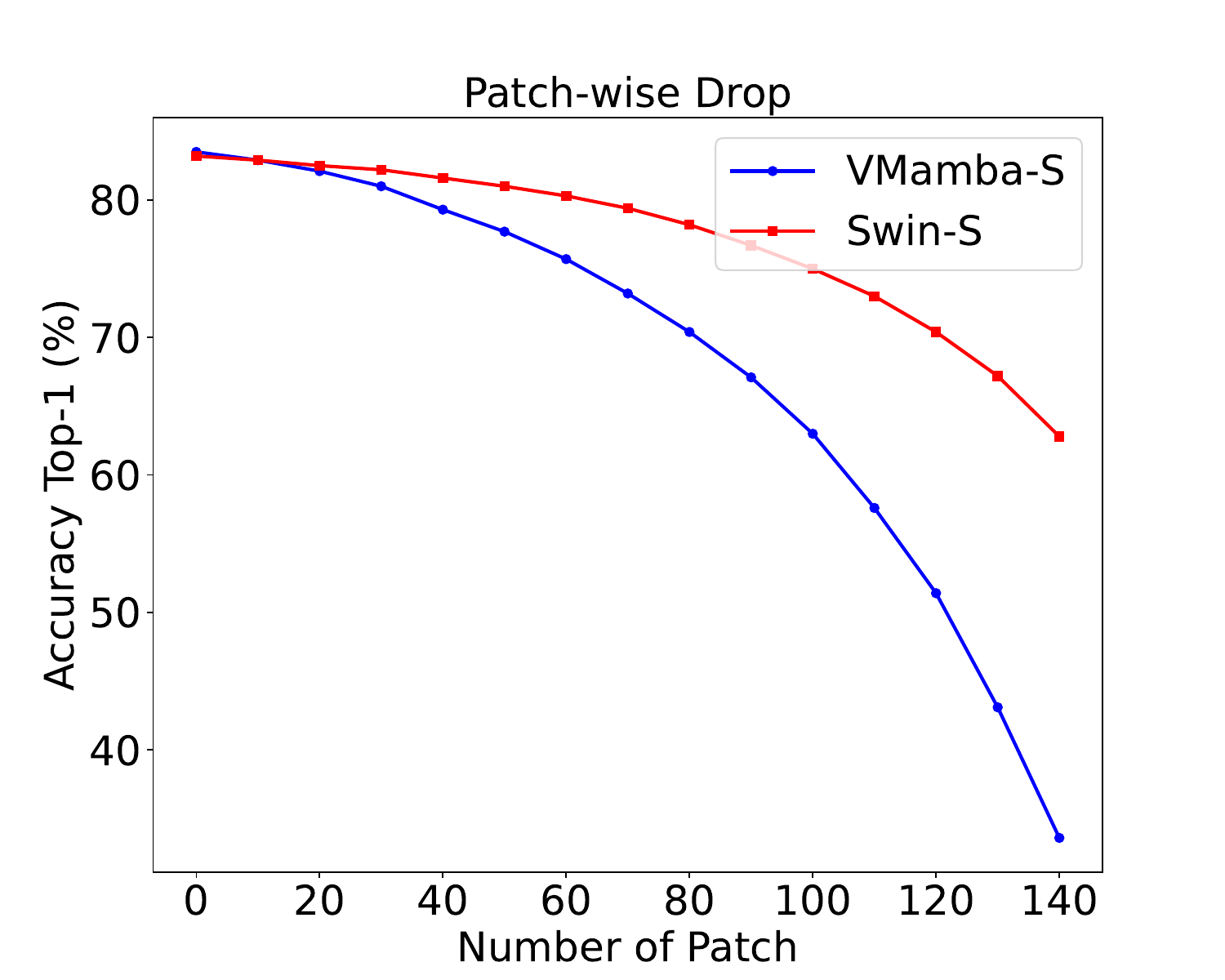}
                \caption{Small}
                \label{fig:RandomDrop_S}
            \end{subfigure}%
            \hfill
            \begin{subfigure}[b]{0.245\textwidth}
                \centering
                \includegraphics[width=0.9\textwidth]{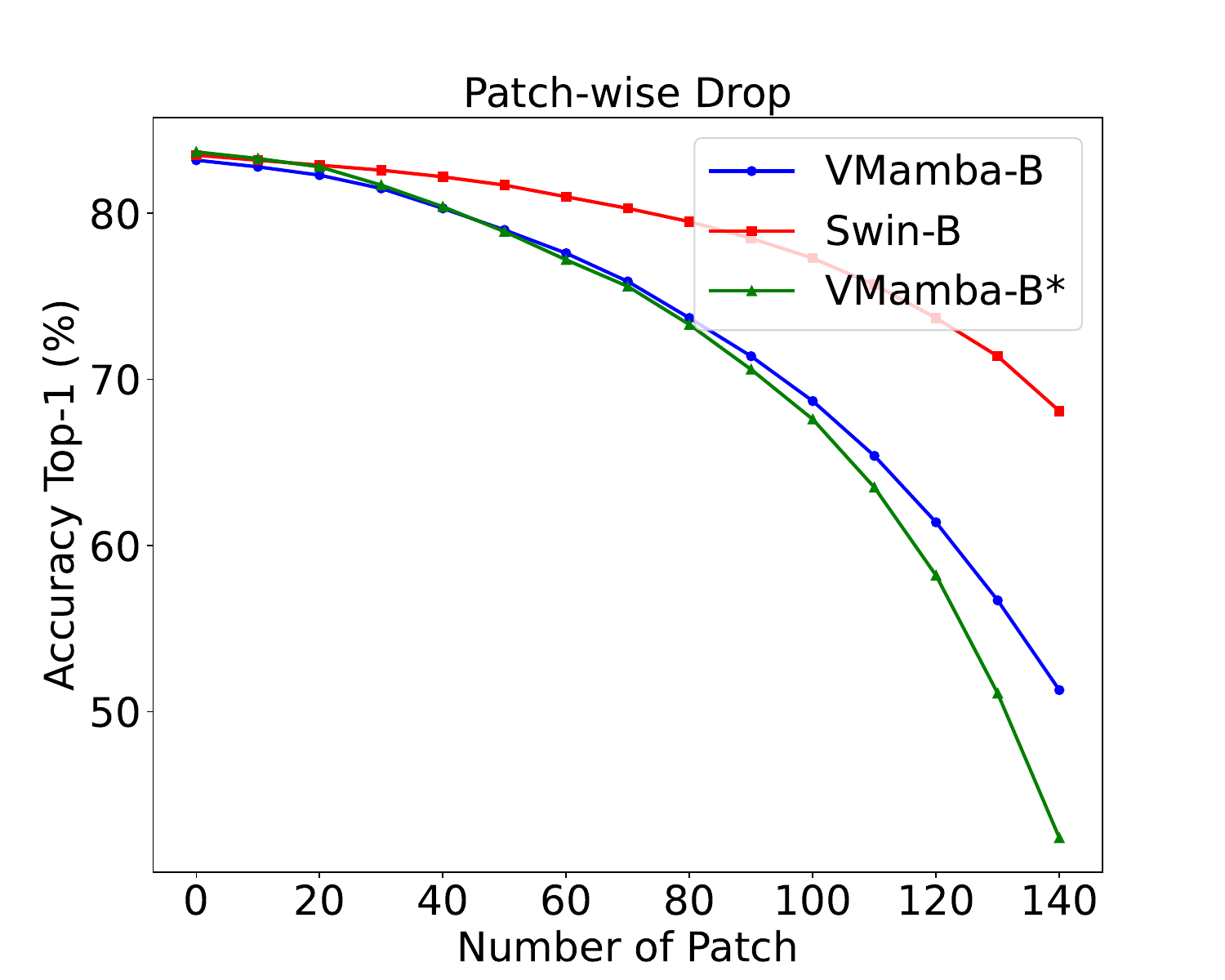}
                \caption{Base}
                \label{fig:RandomDrop_B}
            \end{subfigure}%
            \hfill
            \begin{subfigure}[b]{0.245\textwidth}
                \centering
                \includegraphics[width=0.9\textwidth]{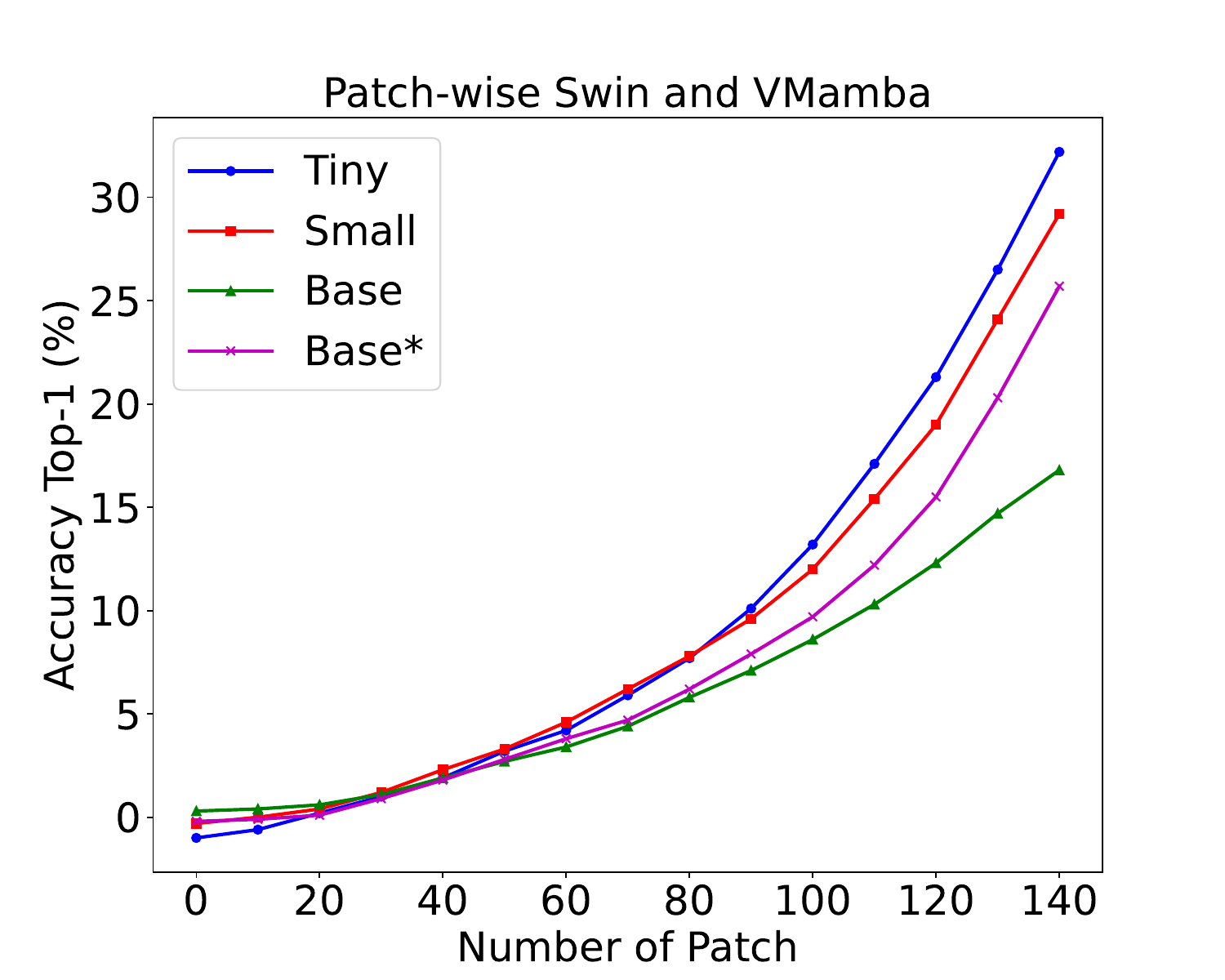}
                \caption{Accuracy Difference}
                \label{fig:Rand_accuracy_differences}
            \end{subfigure}%

            \begin{subfigure}[b]{0.245\textwidth}
                \centering
                \includegraphics[width=0.9\textwidth]{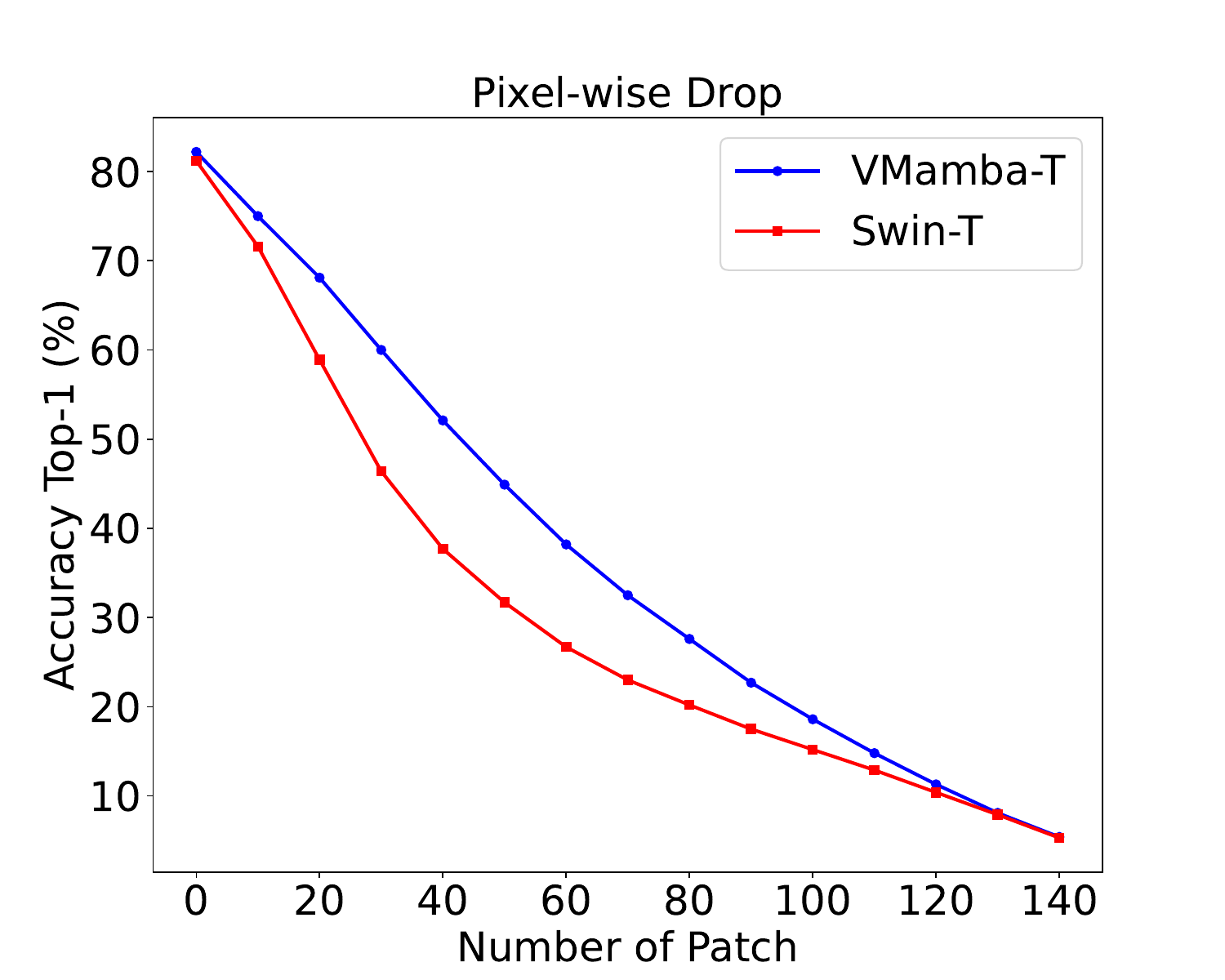}
                \caption{Tiny}
                \label{fig:Pixel_T}
            \end{subfigure}%
            \hfill
            \begin{subfigure}[b]{0.245\textwidth}
                \centering
                \includegraphics[width=0.9\textwidth]{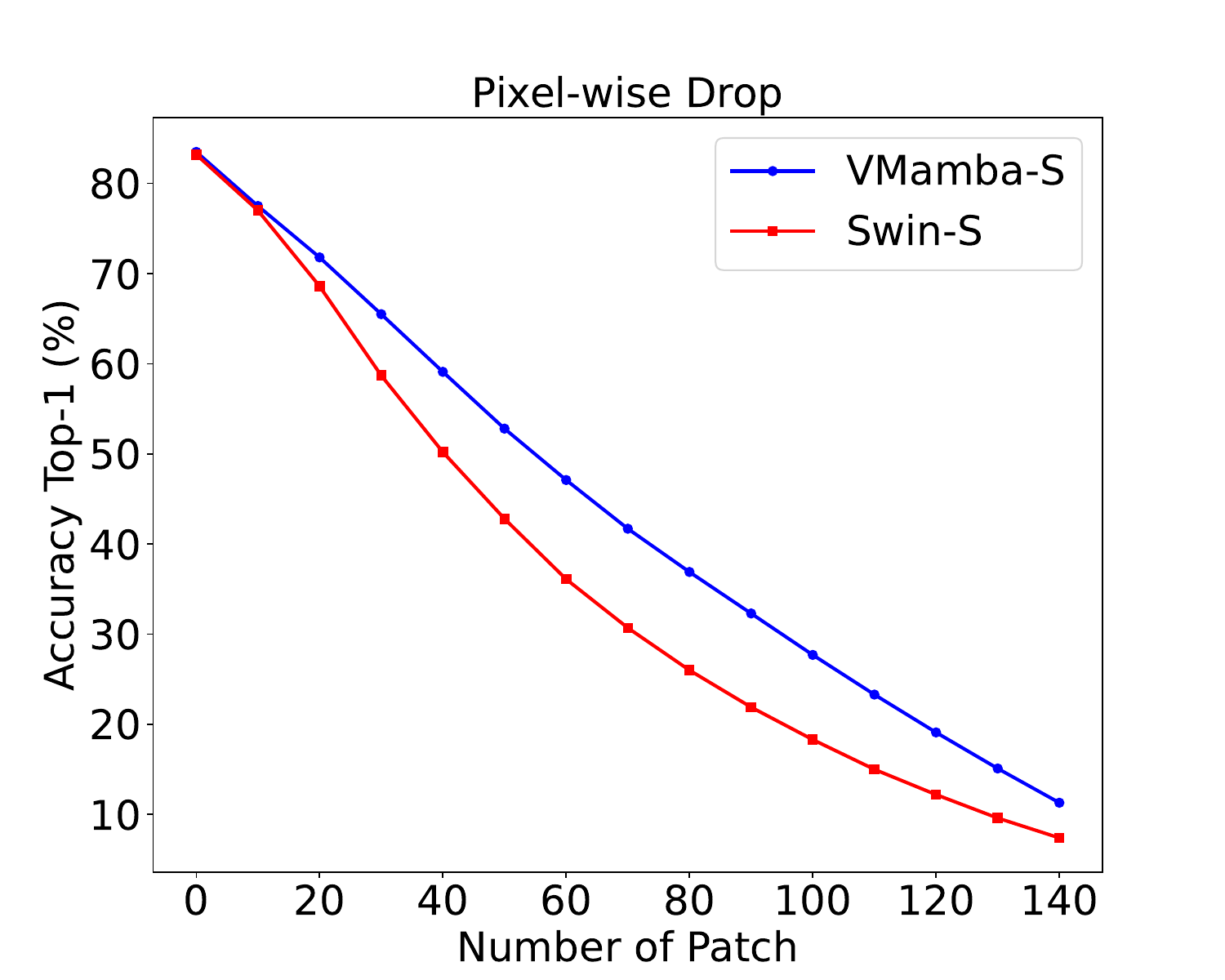}
                \caption{Small}
                \label{fig:Pixel_S}
            \end{subfigure}%
            \hfill
            \begin{subfigure}[b]{0.245\textwidth}
                \centering
                \includegraphics[width=0.9\textwidth]{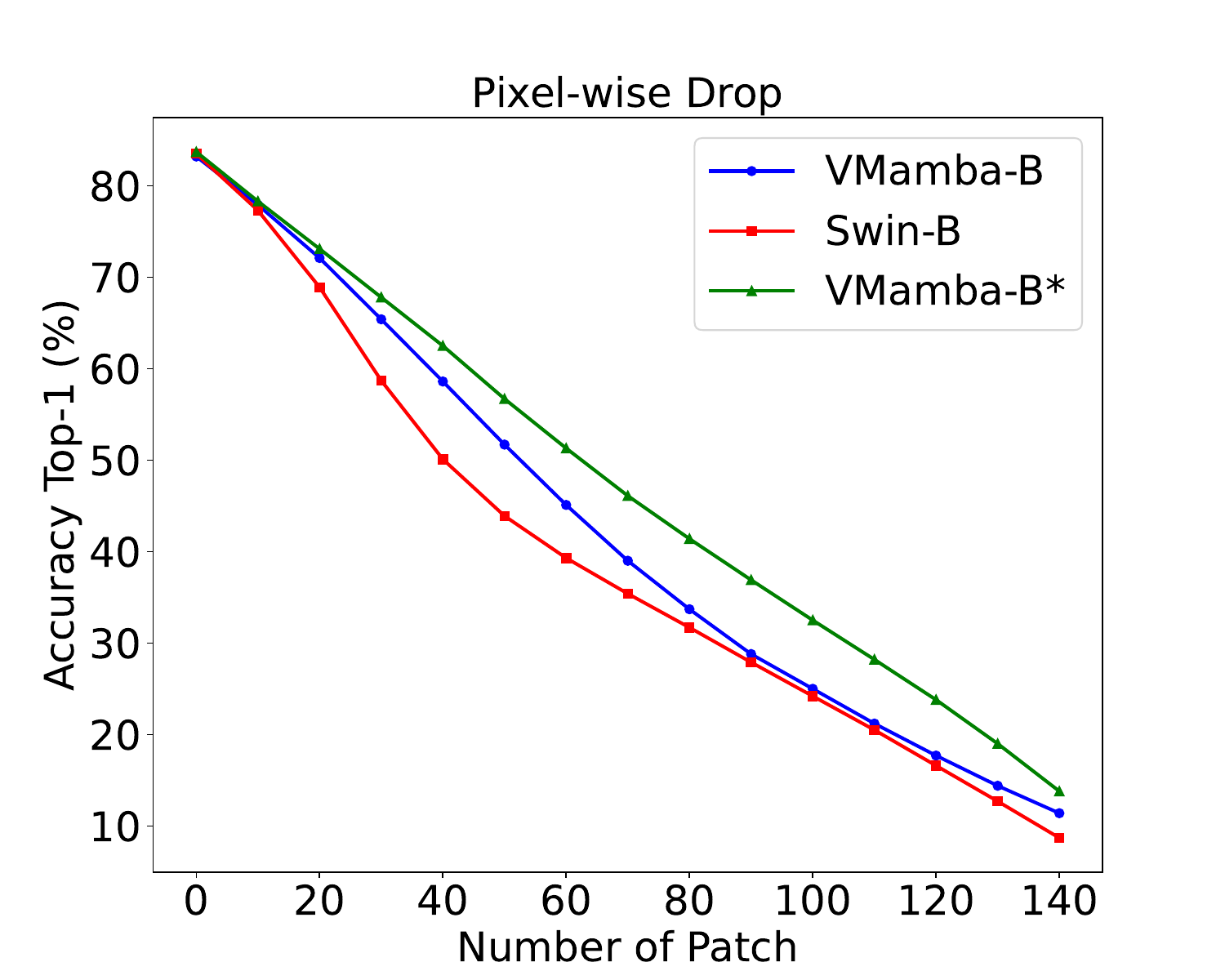}
                \caption{Base}
                \label{fig:Pixel_B}
            \end{subfigure}%
            \hfill
            \begin{subfigure}[b]{0.245\textwidth}
                \centering
                \includegraphics[width=0.9\textwidth]{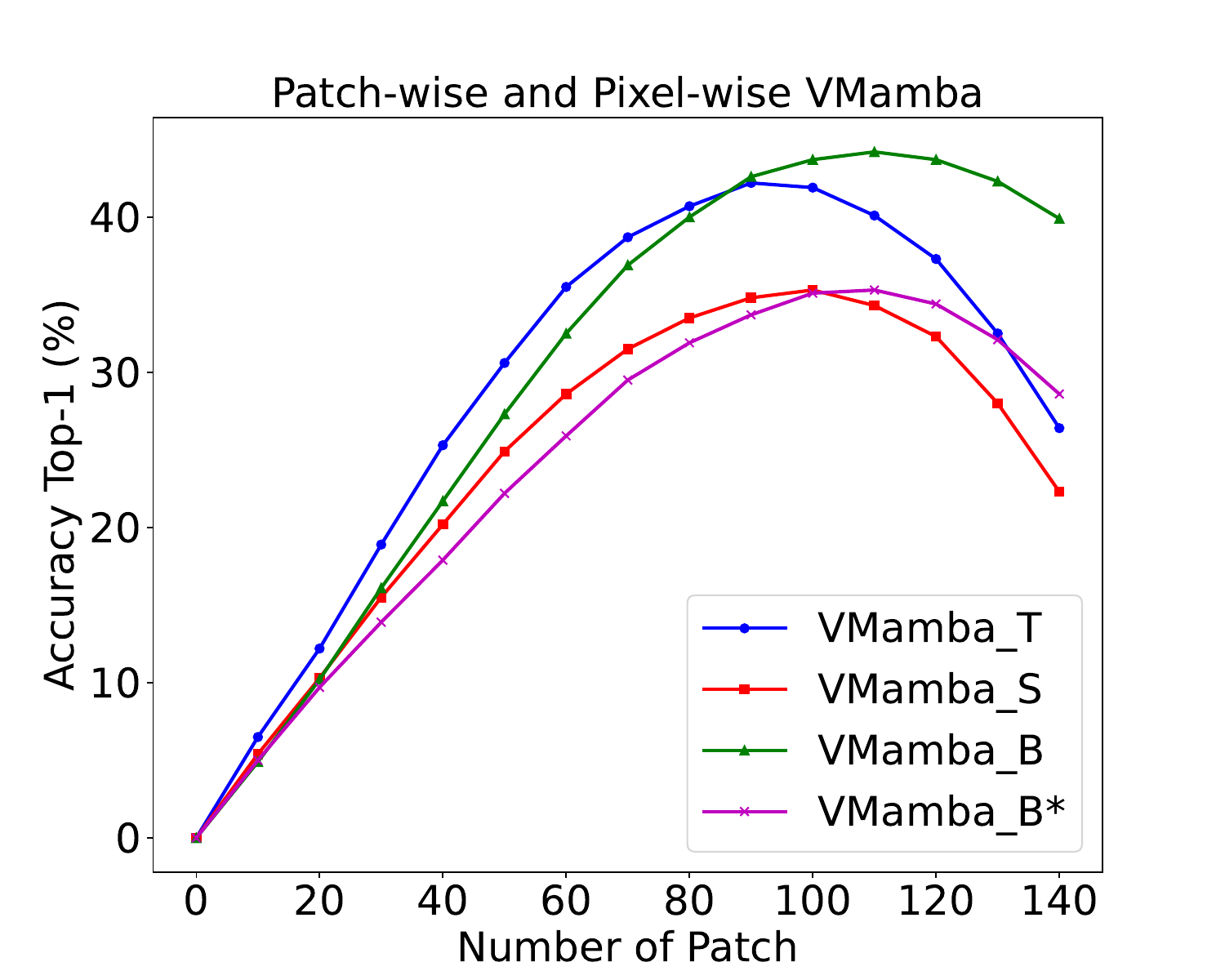}
                \caption{Accuracy Difference}
                \label{fig:Pixel_diff}
            \end{subfigure}%
            \caption{Robust Accuracy related to the amount of information loss.}
            \label{fig:occlusions_exp}
        \end{figure}

        The Transformer and VMamba models both require converting images into sequences of patches of a specified length $L$ to predict labels $Y$. The Transformer models utilize self-attention to facilitate parallel interactions between patches. In contrast, the VMamba model introduces an innovative 2D selective scanning method to process images in both horizontal and vertical directions. 
        
        To analyze the impact of the patch interaction mechanisms adopted by the two models on model robustness, we design two different experiments from the perspectives of both dense and sparse perturbations. \textit{Dense perturbation} means destroying all information within a single patch, while \textit{sparse perturbation} means distributing an equal amount of perturbation across multiple patches. The main difference between these two methods lies in the granularity of the information omission. Patch-wise drop affects larger, contiguous areas of the image, while the pixel-wise drop is employed to evenly distribute perturbations across each patch, which means the perturbation is more sparse and covers a wider range.
        
        The experimental results are shown in the Fig.\ref{fig:occlusions_exp}. The horizontal axis (number of patches) quantifies the amount of information equivalent to how many patches were lost, for example, dropping 10 patches of size 16$\times$16 is equivalent to 256$\times$10 pixels, and the vertical axis represents robust accuracy. In this section, the drop (or loss) of patches or pixels means setting the corresponding values to zero.        

        \subsection{Sensitivity to the Dense Perturbation}
        
        The dense perturbation, i.e. patch-wise drops, randomly selects and drops from the 196 patches (Fig. \ref{fig:patch_case}). In such a setup, the robust performance of Swin is far superior to that of VMamba across all model variants as demonstrated in the first row of Fig. \ref{fig:occlusions_exp}. Notably, the trend of a significant performance decline in VMamba variants becomes especially apparent as the number of patches increases. This phenomenon indicates that:
        \begin{itemize}
            \item[$\bullet$] \textbf{VMamba models require continuity in the scanning trajectory and are sensitive to disturbances along it.}
        \end{itemize}
        There are two possible reasons for this phenomenon: Firstly, the patch-wise drop destroys the structure of the image to some extent, which affects both the Swin and VMamba models. This reason will be further inspected in Section \ref{sec:reletive_position}.
        Secondly, it may disrupt the continuity of the scanning trajectory of the VMamba model.
        
        Specifically, delving into Eq. \ref{eq:Cumulative}, the dependency relationship between patches is sequential and cumulative. This means that the information contained within a current state space is partially contingent upon the preceding patches, forming a chain of dependencies. 
        The patch-wise drop can disrupt this sequential data flow, and when the model tries to make predictions based on incomplete trajectories, it may have a cascading effect on the model's accuracy. However, this sensitivity is reduced to some extent with the increase in model size. Larger VMamba models, such as VMamba-B and VMamba-B*, demonstrate a relative moderation in performance degradation due to patch-wise drop, as shown in Fig. \ref{fig:Rand_accuracy_differences}. This suggests the robustness to dense perturbations within the VMamba framework is scalable, and the increase in model size provides a certain degree of compensatory robustness, enabling these larger models to better withstand the impact of the patch-wise drop.
        
        \subsection{Sensitivity to Sparse Perturbation}

        In the sparse perturbation, we introduced pixel-wise drop which employs a similar randomization strategy to a patch-wise drop but is applied to the individual pixels across the 224x224 pixel space of the image (Fig. \ref{fig:pixel_case}). 
        In the second row of Fig. \ref{fig:occlusions_exp}, we observe that, compared to the Swin model, the VMamba model exhibits greater robustness to pixel-wise perturbations.
        Combined with the experimental results of patch-wise drop, this phenomenon indicates that:
        \begin{itemize}
            \item[$\bullet$] \textbf{Vmamba model has a broader receptive field than Swin model.}
        \end{itemize}
        This makes VMamba more inclined to average the information contained in all patches. Conversely, Swin can effectively utilize local information to capture image features.

        \subsection{Comparison of Dense and Sparse}
    
        Notably, as shown in Fig. \ref{fig:Pixel_diff}, with the increase in the number of patches, the difference in robust accuracy between dense and sparse perturbation shows a sharply rising trend across all variants of the VMamba model, peaking at around 100 patches, with the difference in the range of about 30-40\%.
        This phenomenon indicates that VMamba is affected by error accumulation; that is:
        \begin{itemize}
            \item[$\bullet$] \textbf{With an equal amount of perturbation, the more patches that are affected, the more vulnerable the VMamba model becomes.}
        \end{itemize}
        This meets the expected, as can be seen from Equation \ref{eq:discretization}, where the update mechanism of the intermediate latent state $h(t)$ interacts with each patch along the scanning trajectory, and the small errors produced at each step will accumulate, causing it to eventually deviate from the original prediction result. Besides VMamba-B, which is mentioned in the original paper \cite{liu2024vmamba} as a prototype, the overall trend indicates that as the model size increases, the cumulative error tends to decrease. This phenomenon suggests that improving scalability might be an effective method to address the issue of error accumulation.

    \section{Sensitivity to the Relative Position and Absolute Position of Patches}
 
         We analyze the robustness of VMamba to perturbations in image structure, considering both the relative and absolute positions of patches.
         In terms of \textit{relative position}, the relative order of patch sequence represents the overall image structure and global composition. Concerning learning the contextual information within the patch sequence, Transformer incorporates positional embedding to imbue each patch with spatial information, whereas VMamba discerns this contextual information via the sequential order of its scanning trajectory. 
         In terms of \textit{absolute position}, the consideration of a patch's exact location in the image's spatial layout is equally crucial. This fixed position offers vital spatial cues necessary for a comprehensive grasp of the image as a cohesive entity. A patch positioned in the upper left corner inherently carries a different contextual weight compared to one from the center, as its spatial context can drastically alter its role and the information it conveys when synthesizing the overall image.
    
        \subsection{Relative Position}
        \label{sec:reletive_position}

        \begin{figure}[!tbp]
            \centering
            \begin{subfigure}[b]{0.19\textwidth}
                \centering
                \includegraphics[width=0.9\textwidth]{figures/PatchDrop/image_Rand_0.png}
                \caption{Origin Image}
                \label{fig:origin_shuffle}
            \end{subfigure}%
            \hfill
            \begin{subfigure}[b]{0.19\textwidth}
                \centering
                \includegraphics[width=0.9\textwidth]{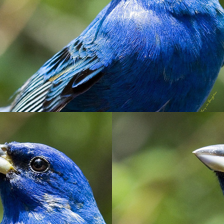}
                \caption{2$\times$2 Grid}
                \label{fig:shuffle_2}
            \end{subfigure}%
            \hfill
            \begin{subfigure}[b]{0.19\textwidth}
                \centering
                \includegraphics[width=0.9\textwidth]{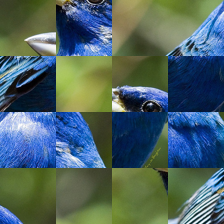}
                \caption{4$\times$4 Grid}
                \label{fig:shuffle_4}
            \end{subfigure}%
            \hfill
            \begin{subfigure}[b]{0.19\textwidth}
                \centering
                \includegraphics[width=0.9\textwidth]{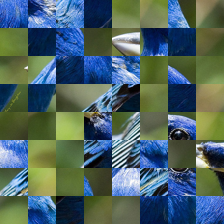}
                \caption{8x8 Grid}
                \label{fig:shuffle_8}
            \end{subfigure}%
            \hfill
            \begin{subfigure}[b]{0.19\textwidth}
                \centering
                \includegraphics[width=0.9\textwidth]{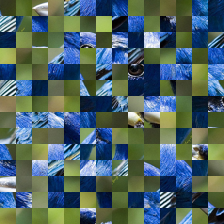}
                \caption{14$\times$14 Grid}
                \label{fig:shuffle_14}
            \end{subfigure}%
            \caption{Example images and their different extend of disorder examples}
            \label{fig:shuffle_case}
        \end{figure}

\begin{figure}[!tbp]
            \centering
            \begin{subfigure}[b]{0.245\textwidth}
                \centering
                \includegraphics[width=0.9\textwidth]{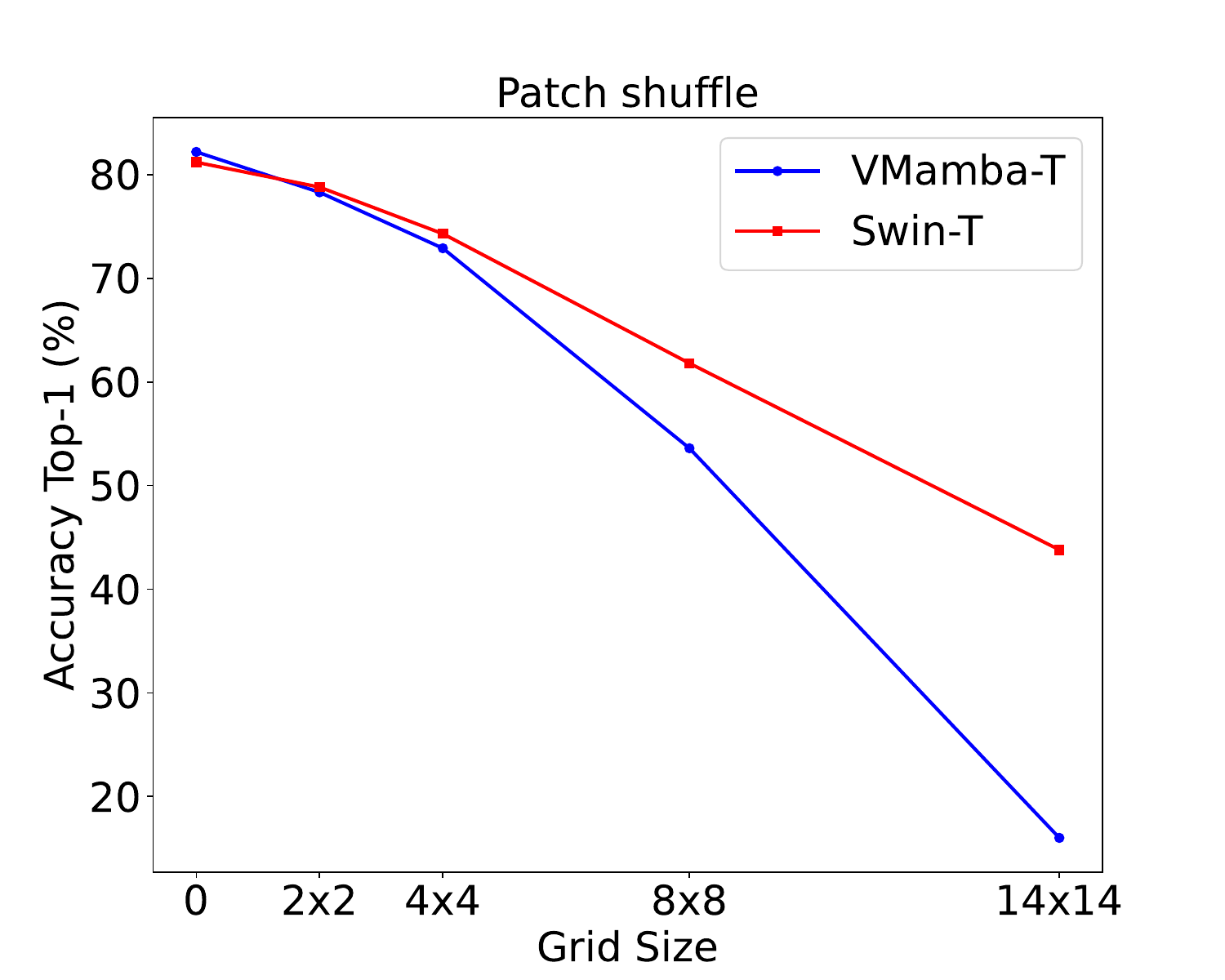}
                \caption{Tiny}
                \label{fig:Origin}
            \end{subfigure}%
            \hfill
            \begin{subfigure}[b]{0.245\textwidth}
                \centering
                \includegraphics[width=0.9\textwidth]{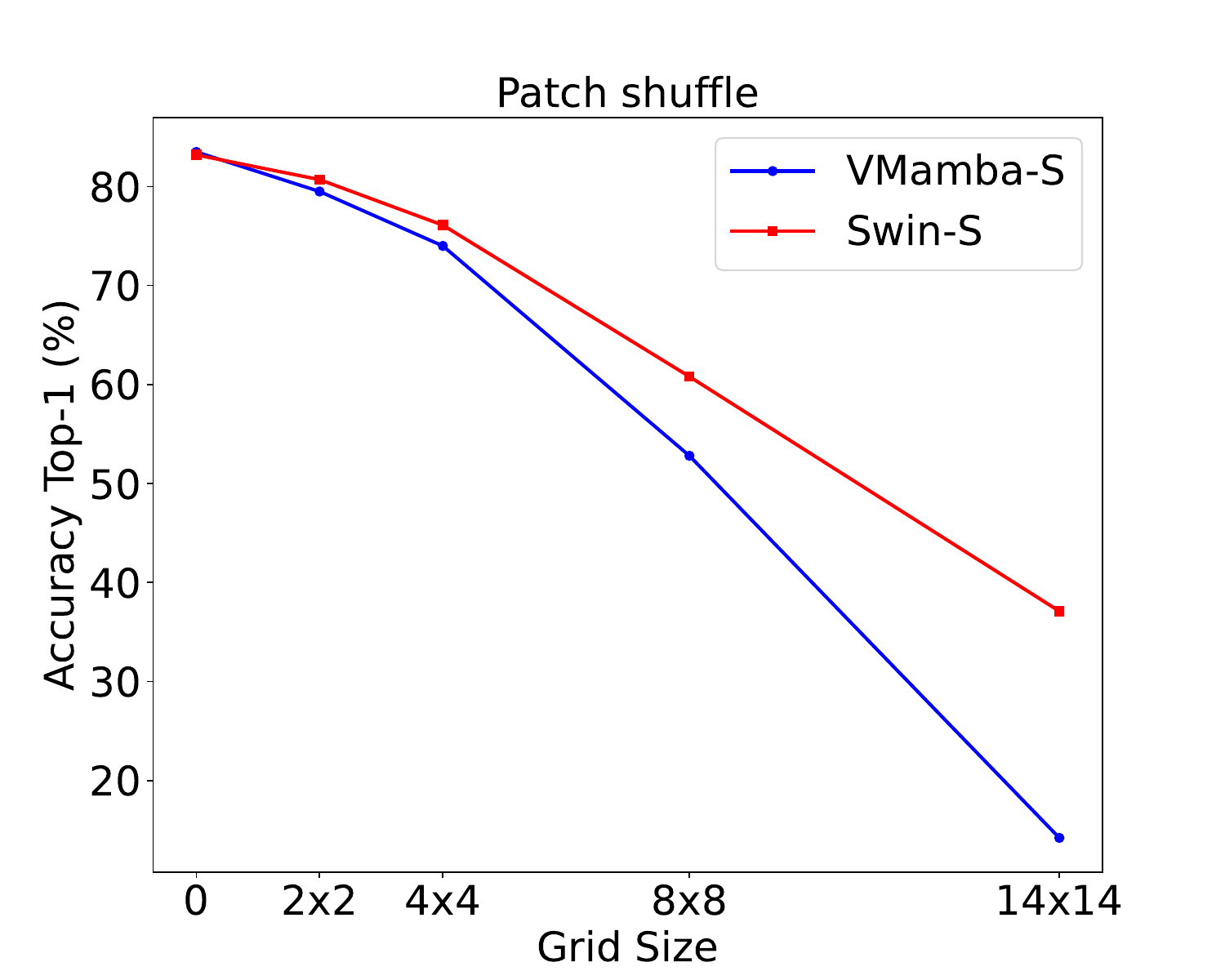}
                \caption{Small}
                \label{fig:shuffle_S}
            \end{subfigure}%
            \hfill
            \begin{subfigure}[b]{0.245\textwidth}
                \centering
                \includegraphics[width=0.9\textwidth]{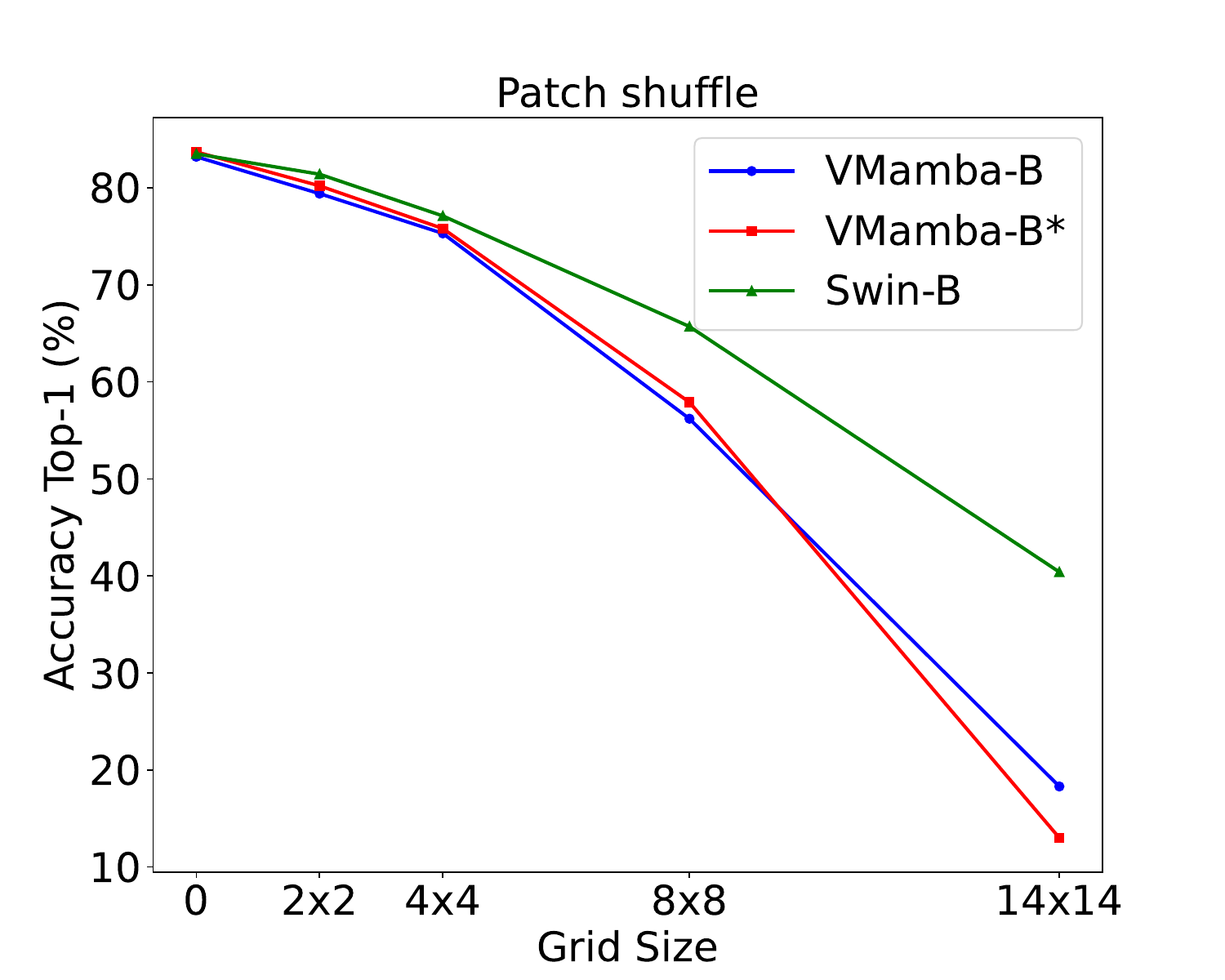}
                \caption{Base}
                \label{fig:shuffle_B}
            \end{subfigure}%
            \hfill
            \begin{subfigure}[b]{0.245\textwidth}
                \centering
                \includegraphics[width=0.9\textwidth]{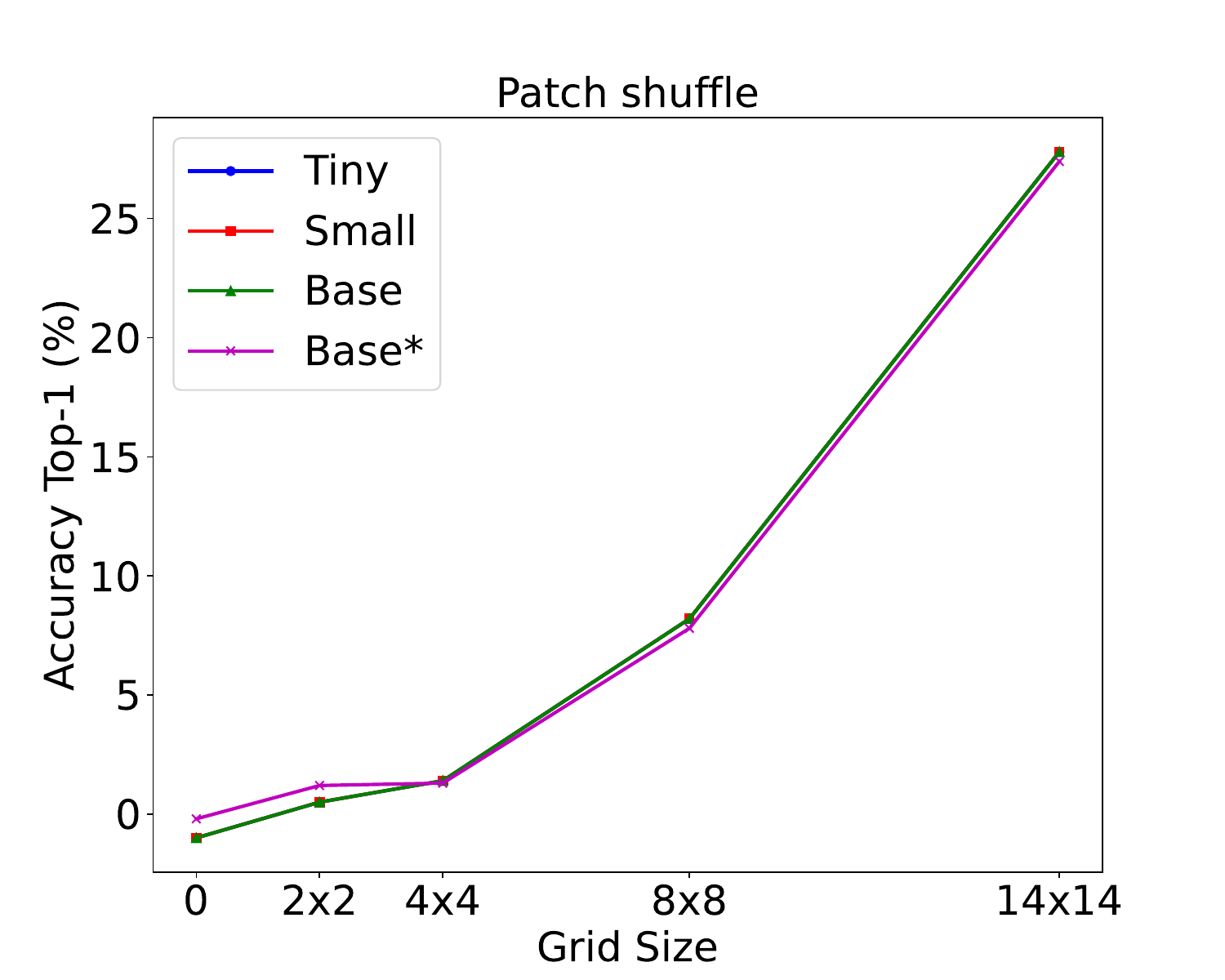}
                \caption{Accuracy Difference}
                \label{fig:shuffle_diff}
            \end{subfigure}%
            \caption{Robust Accuracy related to degrees of shuffle between VMamba and Swin Model.}
            \label{fig:shuffle_exp}
        \end{figure}

         As shown in Fig.\ref{fig:shuffle_case}, when the image is divided into a 2x2 grid, the main subject of the picture can still be easily identified. However, when the grid number is increased to 14x14, it becomes quite challenging to recognize the original subject of the image after the shuffle operation. 
         This indicates:
         \begin{itemize}
            \item[$\bullet$] \textbf{VMamba is highly sensitive to the spatial information of images.}
        \end{itemize}
         Therefore, we will employ the number of grids as the horizontal axis in the following experiments to represent the extent of disorders. Fig.\ref{fig:shuffle_exp} illustrates the correlation between the robust accuracy of the model and the extent of the disorder.
         As the number of grids increases, both VMamba and Swin models exhibit a decline in accuracy, with VMamba showing a more pronounced decrease. For instance, the robust accuracy of VMamba-T and VMamba-S demonstrates a noticeable disparity from that of Swin models following the smallest grid size (2x2), indicating VMamba's heightened sensitivity to the relative order of patches compared to the Swin model. However, the accuracy of VMamba-B and VMamba-B* begins to decline sharply after the 4x4 grid size. This phenomenon may be attributed to the enlarged parameter space of the VMamba architecture, which enables it to more effectively learn the spatial relationships between patches.

    \subsection{Absolute Position}

\begin{figure}[!tbp]
        \centering
        \begin{subfigure}[b]{0.48\textwidth}
            \centering
            \includegraphics[width=0.9\textwidth]{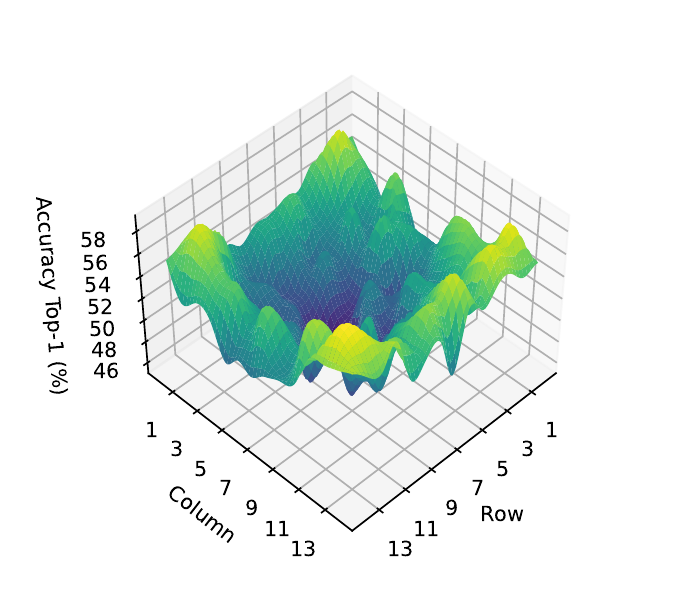}
            \label{fig:heatmap_T}
            \caption{VMamba-T}
        \end{subfigure}%
        \hfill
        \begin{subfigure}[b]{0.48\textwidth}
            \centering
            \includegraphics[width=0.9\textwidth]{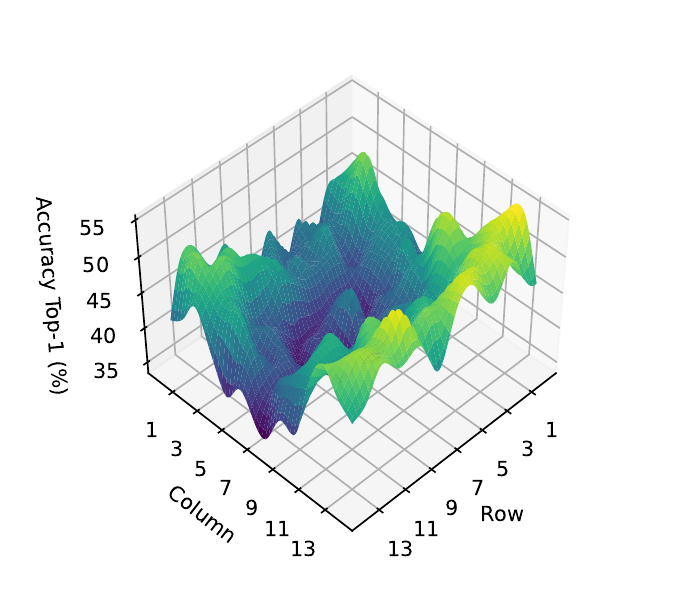}
            \label{fig:heatmap_S}
            \caption{Swin-T}
        \end{subfigure}%
        \caption{Performance of the VMamba-T and Swin-T models under Patch-fool attacks targeting each of 196 image patches, For the efficiency of this experiment, we randomly selected 500 images from the ImageNet-1K validation set.}
        \label{fig:heatmap_exp}
    \end{figure}

    The 3D surface plots Fig. \ref{fig:heatmap_exp} illustrate the performance of the VMamba and Swin models under a white-box adversarial attack targeting the absolute positions of 196 image patches. For the VMamba model, the plot on the left indicates a vulnerability trend where patches near the center are more prone to adversarial attacks, as evidenced by the dips in accuracy within the central region. This central susceptibility is further contextualized by the operational dynamics of the VMamba model, where the image's central region is generally the culmination point within its scanning trajectory.
    This indicates:
    \begin{itemize}
        \item[$\bullet$] \textbf{The closer the perturbation is to the center of the image, the more vulnerable VMamba will be.}
    \end{itemize}
    This insight might suggest that the model's scanning process, which terminally focuses on the central part of the image, could be a contributing factor to the observed vulnerability pattern.
    Contrastingly, the Swin model, as shown on the right, exhibits a more irregular and turbulent response to the adversarial attacks. The sharp fluctuations in accuracy across different patch positions suggest that the Swin model's performance is unevenly affected by the perturbations, with certain areas being more resilient than others. Comparing the two, VMamba's uniformity in performance degradation across patch positions, with a notable central vulnerability, contrasts with Swin's more erratic response, indicating different internal processing and utilization of spatial information within the models. 

\section{Insights}
\label{sec:insights}

    \subsubsection{Address vulnerabilities and defensive capabilities of SSM blocks: } 

    By analyzing the impact of parameters A, B, C, and \(\Delta\) on model robustness, we found that parameters B and C are the main sources of model vulnerability, this vulnerability is especially pronounced in correlation with the size of the model. On the other hand, the \(\Delta\) parameter is highlighted for its defensive capabilities against white-box attacks, with its effectiveness becoming more pronounced as the model size increases. This unique characteristic of the \(\Delta\)  parameter suggests it plays a crucial role in enhancing the model's robustness against such attacks, possibly by disrupting the predictability of the gradient estimation process utilized by adversaries during white-box attacks. Therefore, to enhance VMamba's robustness, future efforts should focus on two primary strategies.
    \begin{itemize}
        \item[$\bullet$] The first strategy involves diminishing the model's output dependency on parameters B and C. This could be achieved through advanced regularization techniques during training, such as dropout, weight decay, or L1/L2 regularization, which would limit these parameters' dominance and mitigate their contribution to the model's vulnerability. The application of these techniques can prevent parameters B and C from becoming focal points of attack, thereby enhancing the overall robustness of the model. 
        \item[$\bullet$] The second strategy revolves around amplifying the defensive capabilities inherent to the \(\Delta\)  parameter. This could involve developing and implementing mechanisms that specifically leverage \(\Delta\) 's protective features, perhaps through targeted training approaches or architectural modifications that emphasize \(\Delta\) 's role in the model's defense against attacks. By doing so, the model can harness the unique protective properties of the \(\Delta\)  parameter more effectively, thereby improving its resilience to adversarial attacks and contributing to a more robust architectural design overall.
    \end{itemize}
    
    \subsubsection{Investigate alternative scanning strategies: } Unlike Transformer architectures that utilize positional embeddings, VMamba's scanning strategy renders it highly sensitive to the structure of images. This heightened sensitivity might be attributed to the training phase where spatially proximate patches tend to be scanned in close succession. However, attempting to mitigate the model's sensitivity to image structure by exhaustively enumerating all potential scanning paths is computationally infeasible. This impracticality stems from the combinatorial explosion of possible paths, which scales exponentially with the size of the image. Given these constraints, the VMamba model necessitates the development of alternative or supplemental mechanisms that can effectively understand and leverage the complex contextual relationships among patches in an image. Several possible solutions could address this sensitivity to image structure without the need for an exhaustive enumeration of scanning paths.
    \begin{itemize}
        \item[$\bullet$] Adaptive Scanning Mechanisms: Introduce adaptive scanning mechanisms that dynamically determine the scanning path based on the image's content or structure. This approach would allow the model to adjust its scanning strategy in real time, focusing on the most informative parts of the image first. Machine learning techniques such as reinforcement learning could be employed to learn optimal scanning paths for different types of images.
        \item[$\bullet$] Hierarchical Scanning Patterns: Implement hierarchical scanning patterns that abstract the image at different levels before deciding on a scanning path. By first analyzing the image at a higher level, the model could identify regions of interest and prioritize scanning in those areas, thereby reducing the dependency on the exact order of patches.
    \end{itemize}

    \subsubsection{Reduced sensitivity to information loss: } For the two different types of information loss, Patch-wise drop and Pixel-wise drop, VMamba shows different vulnerabilities. Specifically, in the case of Patch-wise drop, the loss of information not only disrupts the overall structure of the image but also breaks the continuity of the model's scanning process, severely affecting the model's ability to understand the input data.
    \begin{itemize}
        \item[$\bullet$] To address this issue, we could develop adaptive scanning strategies that allow the model to dynamically adjust its scanning path in response to detected patch drops. By identifying areas of information loss in the image, the model can reroute its scanning trajectory to prioritize intact areas and infer missing information based on the context and spatial relationships of the remaining patches.
    \end{itemize}
    On the other hand, in the scenario of Pixel-wise drop, the impact on the overall structure of the image is relatively minor and does not completely disrupt the continuity of the scanning process. However, each step of scanning generates minor errors, leading to a significant deviation in the final output from what is expected. 
    \begin{itemize}
        \item[$\bullet$] A potential solution to mitigate this issue involves robust feature extraction techniques that tolerate minor errors. Specifically, during the training process, we can introduce a small perturbation to intermediate latent state $h(t)$ to reduce the model's dependency on the results of the previous scan, thereby enhancing the model's robustness.
    \end{itemize}

\section{Conclusion}
    
    In conclusion, our thorough examination of the VMamba emphasizes its considerable promise in computer vision tasks. While excelling in performance across various domains, our focus on robustness reveals nuanced aspects. VMamba demonstrates superior robustness to adversarial attacks compared to Transformer architectures, yet exposes scalability vulnerabilities.
    General robustness assessments showcase remarkable out-of-distribution generalizability but unveil weaknesses against natural adversarial examples and common corruptions.
    Exploring VMamba's gradients and back-propagation during white-box attacks exposes unique vulnerabilities and defensive capabilities within its novel components.
    Furthermore, sensitivity analysis elucidates vulnerabilities associated with the distribution of disturbance area and spatial information, particularly accentuated near the image center.
    This comprehensive inquiry advances our understanding of VMamba's robustness, offering insights pivotal for the refinement and enhancement of VMamba.


%
%
\bibliographystyle{splncs04}
\bibliography{main}

\newpage
\renewcommand{\thesection}{Appendix \Alph{section}}
\renewcommand{\thesubsection}{\Alph{section}.\arabic{subsection}}

\counterwithin{table}{section}
\counterwithin{figure}{section}
\renewcommand{\thetable}{\Alph{section}.\arabic{table}}
\renewcommand{\thefigure}{\Alph{section}.\arabic{figure}}
\setcounter{section}{0} 

\section{Vision Mamba (Vim)}

        VMamba \cite{liu2024vmamba} and Vim \cite{zhu2024vision} are currently the two mainstream frameworks of visual SSM models. VMamba presents an innovative 2D selective scanning method capable of processing images across horizontal and vertical planes and establishing a hierarchical framework similar to the Swin Transformer \cite{liu2021swin}. In a separate development, Vim employs an architecture similar to ViT but replace the Transformer blocks with bi-directional Mamba blocks.

    \subsection{Black-box Attacks and Transferability of Noises}

        In this section, we comprehensively assessed the transferability of adversarial samples between the Vim and transformer models. In Table \ref{tab:Blackbox-vim}, the first column "A vs. B" indicates the two models used for assessment. The second and third columns respectively show the robust accuracy of the A and B models on clean images. The fourth column, "Left to Right", represents the robust accuracy after transferring the adversarial samples generated using model A's gradients to model B, and vice versa. 
        
\begin{figure}[!tbp]
            \centering
            \begin{subfigure}[b]{0.245\textwidth}
                \centering
                \includegraphics[width=0.8\textwidth]{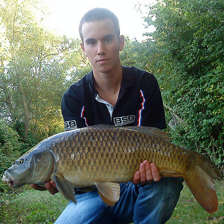}
                \caption{Origin Image}
                \label{fig:Origin_noise}
            \end{subfigure}%
            \hfill
            \begin{subfigure}[b]{0.245\textwidth}
                \centering
                \includegraphics[width=0.8\textwidth]{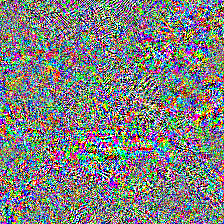}
                \caption{VMamba-T}
                \label{fig:VMamba_noise-1}
            \end{subfigure}%
            \hfill
            \begin{subfigure}[b]{0.245\textwidth}
                \centering
                \includegraphics[width=0.8\textwidth]{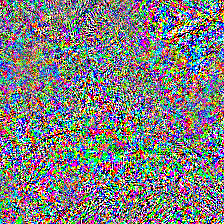}
                \caption{VMamba-S}
                \label{fig:VMamba_noise-2}
            \end{subfigure}%
            \hfill
            \begin{subfigure}[b]{0.245\textwidth}
                \centering
                \includegraphics[width=0.8\textwidth]{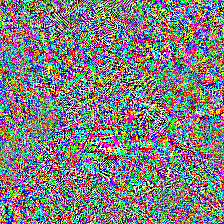}
                \caption{VMamba-B}
                \label{fig:VMamba_noise-3}
            \end{subfigure}%
            \hfill
            \begin{subfigure}[b]{0.245\textwidth}
                \centering
                \includegraphics[width=0.8\textwidth]{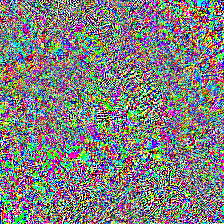}
                \caption{VMamba-B*}
                \label{fig:VMamba_noise-4}
            \end{subfigure}%
            \hfill
            \begin{subfigure}[b]{0.245\textwidth}
                \centering
                \includegraphics[width=0.8\textwidth]{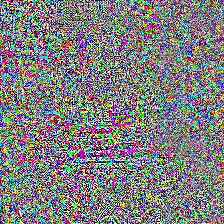}
                \caption{Swin-T}
                \label{fig:Swin_noise-1}
            \end{subfigure}
            \hfill
            \begin{subfigure}[b]{0.245\textwidth}
                \centering
                \includegraphics[width=0.8\textwidth]{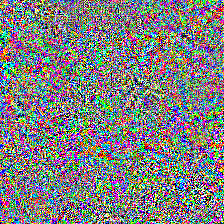}
                \caption{Swin-S}
                \label{fig:Swin_noise-2}
            \end{subfigure}
            \hfill
            \begin{subfigure}[b]{0.245\textwidth}
                \centering
                \includegraphics[width=0.8\textwidth]{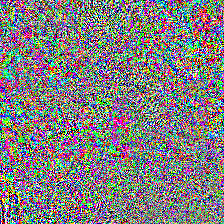}
                \caption{Swin-B}
                \label{fig:Swin_noise-3}
            \end{subfigure}
            \caption{The example perturbations generated with PGD for VMamba and Swin model.}
            \label{fig:noise_case_vm}
        \end{figure}

\begin{figure}[!tbp]
            \centering
            \begin{subfigure}[b]{0.19\textwidth}
                \centering
                \includegraphics[width=0.8\textwidth]{figures/Noise/image_8.png}
                \caption{Origin Image}
                \label{fig:Origin_noise2}
            \end{subfigure}%
            \hfill
            \begin{subfigure}[b]{0.19\textwidth}
                \centering
                \includegraphics[width=0.8\textwidth]{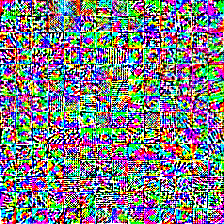}
                \caption{DeiT-Ti}
                \label{fig:DeiT_noiseT}
            \end{subfigure}%
            \hfill
            \begin{subfigure}[b]{0.19\textwidth}
                \centering
                \includegraphics[width=0.8\textwidth]{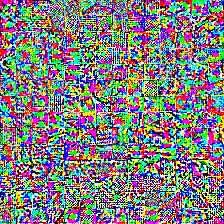}
                \caption{DeiT-S}
                \label{fig:DeiT_noiseS}
            \end{subfigure}%
            \hfill
            \begin{subfigure}[b]{0.19\textwidth}
                \centering
                \includegraphics[width=0.8\textwidth]{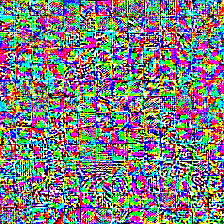}
                \caption{DeiT-B}
                \label{fig:DeiT_noiseB}
            \end{subfigure}%
            \hfill
            \begin{subfigure}[b]{0.19\textwidth}
                \centering
                \includegraphics[width=0.8\textwidth]{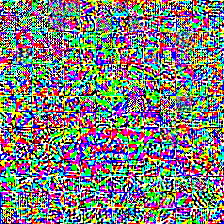}
                \caption{ViT-S/16}
                \label{fig:ViT_noiseS}
            \end{subfigure}%
            \hfill
            \begin{subfigure}[b]{0.19\textwidth}
                \centering
                \includegraphics[width=0.8\textwidth]{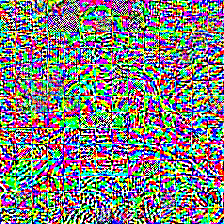}
                \caption{ViT-B/16}
                \label{fig:ViT_noiseB}
            \end{subfigure}%
            \hfill
            \begin{subfigure}[b]{0.19\textwidth}
                \centering
                \includegraphics[width=0.8\textwidth]{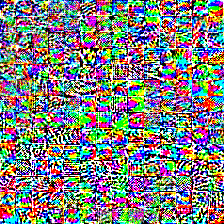}
                \caption{Vim-tiny}
                \label{fig:Swin_noise-4}
            \end{subfigure}
            \hfill
            \begin{subfigure}[b]{0.19\textwidth}
                \centering
                \includegraphics[width=0.8\textwidth]{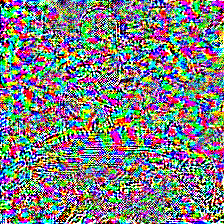}
                \caption{Vim-tiny+}
                \label{fig:Swin_noise-5}
            \end{subfigure}
            \hfill
            \begin{subfigure}[b]{0.19\textwidth}
                \centering
                \includegraphics[width=0.8\textwidth]{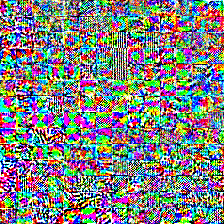}
                \caption{Vim-small}
                \label{fig:Swin_noise-6}
            \end{subfigure}
            \hfill
            \begin{subfigure}[b]{0.19\textwidth}
                \centering
                \includegraphics[width=0.8\textwidth]{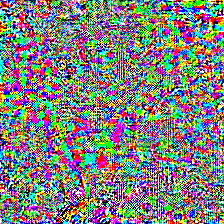}
                \caption{Vim-small+}
                \label{fig:Swin_noise-7}
            \end{subfigure}
            \caption{The example perturbations generated with PGD for Vim, DeiT, and ViT model.}
            \label{fig:noise_case_vim}
        \end{figure}

        From Table \ref{tab:Blackbox-vim}, the Vim model demonstrates the highest transferability of adversarial samples with the DeiT model across all sizes, followed by the ViT model. Therefore, to reasonably compare the robustness differences between Vim and Transformer models, we prefer to compare VMamba with Swin, and Vim with ViT and DeiT. Fig.\ref{fig:noise_case_vm} and Fig. \ref{fig:noise_case_vim} illustrate the example perturbations generated for VMamba, Vim and Transformer models utilizing the PGD method.
        \begin{table}[!tbp]
            \centering
            \caption{Transferability of adversarial samples between the SSM and transformer models. The first column "A vs. B" indicates the two models used for assessment. The second and third columns respectively show the robust accuracy of the A and B models on clean images. The fourth column, "Left to Right", represents the robust accuracy after transferring the adversarial samples generated using model A's gradients to model B, and vice versa. All adversarial samples are generated by using PGD. Vim\textsuperscript{\textdagger} is the version adapted for Long Sequence Fine-tuning. Specifically, they maintained the original patch size but adjusted the patch extraction stride to 8.}
            \label{tab:Blackbox-vim}
            \begin{tabular}{l|cc|c|c}
                \toprule
                    \textbf{} & \textbf{SSM} &
                    \textbf{Transformer} & \textbf{Left $\rightarrow$ Right} & \textbf{Right $\rightarrow$ Left} \\
                \midrule    
                    Vim-Ti vs. DeiT-Ti & 76.1 & 72.2 & 59.4 & \textbf{66.4}   \\
                    Vim-Ti vs. Swin-T & 76.1 & 81.2 & 77.2 & 74.0   \\
                    Vim-Ti\textsuperscript{\textdagger} vs. DeiT-Ti & 78.3 & 72.2 & 63.4 & \textbf{73.5}   \\
                    Vim-Ti\textsuperscript{\textdagger} vs. Swin-T & 78.3 & 81.2 & 76.1 & 76.0   \\
                \midrule   
                    Vim-S vs. DeiT-S & 80.5 & 79.8 & 70.9 & \textbf{70.7}   \\
                    Vim-S vs. ViT-S & 80.5 & 74.7 & 58.3 & 74.8   \\
                    Vim-S vs. Swin-S & 80.5 & 83.2 & 79.3 & 78.2   \\
                    Vim-S\textsuperscript{\textdagger} vs. DeiT-S & 81.6 & 79.8 & 73.4 & \textbf{76.2}   \\
                    Vim-S\textsuperscript{\textdagger} vs. ViT-S & 81.6 & 74.7 & 60.0 & 78.3   \\
                    Vim-S\textsuperscript{\textdagger} vs. Swin-S & 81.6 & 83.2 & 78.6 & 79.2   \\


                \bottomrule
            \end{tabular}
        \end{table}

\subsection{White-box Attacks and General Robustness}

    Tables \ref{tab:exp:main-results-vim} and \ref{tab:Number_patches_vim} delineate the experimental results for the two aforementioned white-box attack settings: "Attack the entire image" and "Patch-wise attack". These tables reveal that the Vim model consistently outperforms the DeiT and ViT models in terms of robust accuracy across various attack scenarios, thereby evidencing its superior adversarial robustness. For instance, as for Attack the entire image, comparing to DeiT-S, Vim-small, and Vim-small\textsuperscript{\textdagger} achieve 12.9\% and 22.3\% higher accuracy under FGSM, and 19.7\% and 25.0\% higher accuracy under PGD, respectively. In the context of the Patch-wise attack scenario, where a singular patch is subjected to adversarial attack, the ViT-S model experiences a catastrophic decline in performance, evidencing a robust accuracy of merely 0.1\%. Similarly, the DeiT-S model demonstrates a significantly compromised resilience, achieving a robust accuracy of only 6.6\%. In stark contrast, the Vim-small and Vim-small\textsuperscript{\textdagger} models exhibit a considerably higher level of adversarial robustness, maintaining a robust accuracy of 63.5\% and 55.2\% respectively.

        \begin{table}[!tbp]
        \centering
        \caption{Evaluation of SOTA methods on ImageNet-1K. 
        The top-1 accuracy is used to assess performance on clean ImageNet-1K and under adversarial attacks (FGSM and PGD).
        All models utilize input dimensions of $224\times224$. Vim\textsuperscript{\textdagger} is the version adapted for Long Sequence Fine-tuning. Specifically, they maintained the original patch size but adjusted the patch extraction stride to 8.}
        \begin{tabular}{c|l|c|cc}
            \toprule
                \textbf{~Categories~} & 
                \textbf{Models} & 
                \textbf{~Clean~} & 
                \textbf{~FGSM~} &
                \textbf{~PGD~} \\
            \midrule
                \multirow{5}{*}{Transformer}
                & ViT-S/16 + AugReg \cite{steiner2021train} & 74.7 & 27.2 & ~9.3 \\ 
                & DeiT-Ti \cite{pmlr-v139-touvron21a}        & 72.2 & 22.3 & ~6.1 \\
                & DeiT-S \cite{pmlr-v139-touvron21a}        & 79.8 & 40.5 & 16.4 \\ 
                & Swin-T \cite{liu2021swin}                 & 81.2 & 33.8 & ~7.30 \\ 
                & Swin-S \cite{liu2021swin}                 & 83.2 & 45.9 & 18.3 \\ 
            \midrule
                \multirow{4}{*}{\begin{tabular}{@{}c@{}}SSM\end{tabular}}
                & Vim-tiny  \cite{zhu2024vision}            & 76.1 & 39.8 & 21.2 \\ 
                & Vim-tiny\textsuperscript{\textdagger}  \cite{zhu2024vision}            & 78.3 & 47.9 & 28.6 \\ 
                & Vim-small  \cite{zhu2024vision}            & 80.5 & 53.4 & 36.1 \\ 
                & Vim-small\textsuperscript{\textdagger} \cite{zhu2024vision}            & 81.6 & 62.8 & 41.4 \\ 
            \bottomrule
        \end{tabular}
        \label{tab:exp:main-results-vim}
        \end{table}

        \begin{table}[!tbp]
        \centering
        \caption{Robust Accuracy under Patch-fool attack, "P1" to "P4" represent the robust accuracy when a certain number of patches are under attack. P1-P4 signifies the difference in robust accuracy between P1 and P4. Vim\textsuperscript{\textdagger} is the version adapted for Long Sequence Fine-tuning. Specifically, they maintained the original patch size but adjusted the patch extraction stride to 8.}
        \label{tab:Number_patches_vim}
        \begin{tabular}{c|l|c|cccc|c}
            \toprule
                \textbf{~Categories~} &
                \textbf{Model} &
                \textbf{~Clean~} &
                \textbf{~P1~} &
                \textbf{~P2~} &
                \textbf{~P3~} &
                \textbf{~P4~} &
                \textbf{~P1$-$P4~} \\

            \midrule
                \multirow{5}{*}{Transformer}
                & ViT-S/16 + AugReg \cite{steiner2021train} & 75.9 & ~0.1 & ~0.0 & ~0.0 & ~0.0 & ~0.1 \\
                & DeiT-T \cite{pmlr-v139-touvron21a}        & 72.6 & ~4.9 & ~0.0 & ~0.0 & ~0.0 & ~4.9 \\
                & DeiT-S \cite{pmlr-v139-touvron21a}        & 81.4 & ~6.6 & ~0.2 & ~0.0 & ~0.0 & ~6.6 \\
                & Swin-T \cite{liu2021swin}                 & 81.5 & 40.3 & 16.6 & ~5.7 & ~2.2 & 38.1 \\
                & Swin-S \cite{liu2021swin}                 & 84.0 & 52.6 & 22.9 & 10.7 & ~6.0 & 46.6 \\
            \midrule
                \multirow{4}{*}{\begin{tabular}{@{}c@{}}SSM\end{tabular}}
                & Vim-tiny  \cite{zhu2024vision} & 74.8 & 44.1 & 25.6 & 15.5 & 11.5 & 32.6 \\
                & Vim-tiny\textsuperscript{\textdagger}  \cite{zhu2024vision} & 76.7 & 47.0 & 26.6 & 16.0 & 11.6 & 35.4 \\
                & Vim-small  \cite{zhu2024vision} & 84.7 & 63.5 & 45.1 & 29.6 & 21.0 & 42.6 \\
                & Vim-small\textsuperscript{\textdagger} \cite{zhu2024vision} & 86.0 & 55.2 & 28.2 & 16.5 & 9.6 & 45.6 \\
            \bottomrule
        \end{tabular}
        \end{table}

    Table \ref{tab:exp:main-results-variants-vim} reports the experiment results on the dataset ImageNet-A, ImageNet-R, and ImageNet-C.
    %
    %
    Vim showcases remarkable robustness against common corruptions, as evidenced by its significantly lower mCE on the ImageNet-C dataset. Both Vim-small\textsuperscript{\textdagger} and Vim-tiny\textsuperscript{\textdagger} outperform the best Transformer model, Swin, by 3.5\% and 1.9\%, respectively.
    Despite yielding slightly lower accuracy than Swin on ImageNet-A and R datasets, Vim maintains its superiority over ViT and DeiT. This implies that while Vim may not excel in handling natural adversarial examples present in ImageNet-A or the out-of-distribution data of ImageNet-R as effectively as Swin, it still showcases some performance improvements over previous models.
    Interestingly, Vim demonstrates a distinct behavior from VMamba; while VMamba excels in robustness against out-of-distribution data, Vim shines in robustness against common corruptions.
    The scalability concern appears to be absent for Vim when assessed on ImageNet-C. However, Vim only offers tiny and small versions, which restricts the thorough evaluation of its scalability.

        \begin{table}[!tbp]
        \centering
        \caption{Evaluation of SOTA methods on ImageNet variants (A, R, and C). 
        The top-1 accuracy is used to assess performance on ImageNet-A, and -R. In the case of ImageNet-C, the focus is on the mean Corruption Error (mCE), where lower values indicate better performance (marked by ${\downarrow}$). 
        All models utilize input dimensions of $224\times224$. Vim\textsuperscript{\textdagger} is the version adapted for Long Sequence Fine-tuning. Specifically, they maintained the original patch size but adjusted the patch extraction stride to 8.}
        \begin{tabular}{c|l|ccc}
            \toprule
                \textbf{~Categories~} &
                \textbf{Models} &
                \textbf{~~A~~} & \textbf{~~R~~} & \textbf{~C} ${(\downarrow)}$ \\

            \midrule
                \multirow{5}{*}{Transformer}
                & ViT-S/16 + AugReg \cite{steiner2021train} & ~9.0  & 31.9 & 53.4 \\
                & DeiT-T    \cite{pmlr-v139-touvron21a}     & 7.6 & 32.7 & 53.6 \\
                & DeiT-S    \cite{pmlr-v139-touvron21a}     & 19.5 & 41.9 & 41.2 \\
                & Swin-T    \cite{liu2021swin}              & 21.2 & 41.2 & 45.9 \\
                & Swin-S    \cite{liu2021swin}              & 32.6 & 44.8 & 41.0 \\
            \midrule
                \multirow{4}{*}{\begin{tabular}{@{}c@{}}SSM\end{tabular}}
                & Vim-tiny     \cite{zhu2024vision}         & ~9.5 & 38.8 & 46.9 \\
                & Vim-tiny\textsuperscript{\textdagger}     \cite{zhu2024vision}         & 17.2 & 39.7 & 44.0 \\
                & Vim-small     \cite{zhu2024vision}         & 19.7 & 44.7 & 38.9 \\
                & Vim-small\textsuperscript{\textdagger}    \cite{zhu2024vision}         & 28.3 & 44.3 & 37.5 \\
            \bottomrule
        \end{tabular}
        \label{tab:exp:main-results-variants-vim}
        \end{table}

\section{Related Works}
\label{sec:related}

    \subsection{State Space Models}

        State Space Models (SSMs) have emerged in deep learning, demonstrating their effectiveness in efficient long sequence modeling. This success has garnered significant attention from both the Natural Language Processing and Computer Vision communities.
        The Linear State-Space Layer (LSSL) \cite{gu2021combining} combines recurrent, convolutional, and continuous-time models to address their individual shortcomings. The LSSL model demonstrates state-of-the-art performance in time-series tasks, surpassing previous approaches on sequential image classification, healthcare regression, and speech tasks.
        The Structured State Space sequence model (S4) \cite{gu2021efficiently} focuses on efficient modeling of long sequences by optimizing the fundamental SSM. This approach demonstrates strong empirical results across diverse benchmarks, achieving competitive accuracy on sequential CIFAR-10 and outperforming prior methods on the Long Range Arena benchmark, including the challenging Path-X task.
        The S5 model \cite{smith2022simplified} extends the structured state space paradigm with the S5 layer, which leverages multi-input, multi-output SSMs for efficient parallel processing. The S5 layer achieves state-of-the-art results on long-range sequence modeling tasks, showcasing its prowess in tasks like the Long Range Arena benchmark's Path-X.
        Mamba \cite{gu2023mamba} is a sequence model that leverages structured state spaces (SSMs) to achieve linear-time sequence modeling. Notably, Mamba surpasses the computational efficiency of Transformers with a 5$\times$ higher throughput, excelling across modalities such as language, audio, and genomics.

        The S4ND \cite{nguyen2022s4nd} model extends the continuous-signal modeling prowess of state space models (SSMs) to multidimensional data like images and videos. S4ND excels in modeling large-scale visual data in 1D, 2D, and 3D as continuous multidimensional signals, showcasing superior performance on practical tasks. When integrated into existing state-of-the-art models by replacing Conv2D and self-attention layers, S4ND outperforms a Vision Transformer baseline on ImageNet-1k and matches ConvNeXt in 2D image modeling. For video tasks, S4ND improves activity classification on HMDB-51 compared to an inflated 3D ConvNeXt.
        VMamba \cite{liu2024vmamba} introduces a Visual State Space Model inspired by state space models and designed to achieve linear complexity while preserving global receptive fields. VMamba addresses direction-sensitive issues with the Cross-Scan Module (CSM) and exhibits promising capabilities across various visual perception tasks, outperforming established benchmarks as image resolution increases. These works collectively advance the understanding and efficiency of visual representation learning models.

    \subsection{Adversarial Robustness}
    
        Szegedy \textit{et al.} \cite{szegedy2013intriguing} uncovered a significant vulnerability in state-of-the-art neural networks and machine learning models. Their discovery highlighted the vulnerability of these models to adversarial examples. Adversarial examples are instances that lead to misclassifications when they are slightly altered.
        Building upon the work of Szegedy \textit{et al.} \cite{szegedy2013intriguing}, numerous novel methods have been developed to generate adversarial noises, enabling the effective alteration of inputs to models.

        \paragraph{Fast Gradient Sign Method} attack \cite{goodfellow2014explaining} (FGSM) has proven that the linear behavior in high-dimensional spaces is adequate to induce adversarial examples, marking a crucial insight in the realm of adversarial machine learning. This perspective has facilitated the development of a rapid adversarial example generation method, thereby rendering adversarial training more practical.
        Let $\bm{\theta}$ denote the parameters of a model, $\bm{x}$ denote the input, $y$ denote the targets, and $J(\bm{\theta}, \bm{x}, y)$ denote the loss function for training the neural network. The optimal max-norm constrained perturbation, denoted as $\bm{\eta}$ can be calculated by:
        \begin{equation}
            \bm{\eta} = \varepsilon \operatorname{sign} \left( \nabla_{\bm{x}} J(\bm{\theta}, \bm{x}, y)\right),
        \end{equation}
        where $\varepsilon$ is a step size. Then, the adversarial example $\bm{x}' = \bm{x} + \bm{\eta}$ is obtained by adding the perturbation $\bm{\eta}$ to the original input $\bm{x}$.

        \paragraph{Projected Gradient Descent} attack \cite{madry2017towards} (PGD) offers a distinctive perspective on the adversarial attack and defense problem by framing it as a saddle point problem
        \begin{equation}
            \min_{\bm{\theta}} \rho(\bm{\theta}), \quad \mathrm{where~} \rho(\bm{\theta}) = \mathbb{E}_{(\bm{x}, y) \sim \mathcal{D}} \left[ \max_{\bm{\delta} \in \mathcal{S}} J(\bm{\theta}, \bm{x} + \bm{\delta}, y)\right].
        \end{equation}
        This formulation allows PGD to interpret the FGSM attack as a straightforward one-step scheme aimed at maximizing the inner part of the saddle point formulation.
        A more powerful adversary is introduced through the multi-step variant, essentially aligning with the principles of Projected Gradient Descent applied to the negative loss function:
        \begin{equation}
            \bm{x}^{t+1} = \prod_{\bm{x} + \mathcal{S}} \left( \bm{x}^{t} + \varepsilon \operatorname{sign} \left( \nabla_{\bm{x}} J(\bm{\theta}, \bm{x}, y) \right)\right).
        \end{equation}
        This approach broadens the understanding of adversarial attacks, providing a novel view that extends beyond the simple one-step scheme.

        \paragraph{Patch-Fool} attack \cite{fu2022patch} introduces a novel strategy by constraining perturbed pixels within one patch or several patches, unlike previous approaches that limit perturbation strength onto each pixel.
        This method can be viewed as a variant of sparse attacks \cite{modas2019sparsefool, croce2019sparse, dong2020greedyfool}. This approach produces adversarial examples with noisy patches, visually resembling and emulating natural corruptions within a small region of the original image.
        The objective of Patch-Fool can be formulated as:
        \begin{equation}
            \argmax_{1 \leq p \leq n, \bm{E} \in \mathbb{R}^{n \times d}} J \left( \bm{x} + \mathbbm{1}_{p} \odot \bm{E}, y \right)
        \end{equation}
        where $\bm{E}$ is the adversarial perturbation, $\mathbbm{1}_{p} \in \mathbb{R}^{n}$ is a one-hot vector, and $\odot$ represents the penetrating face product.

    \subsection{General Robustness}

        \paragraph{ImageNet-A} \cite{hendrycks2021nae} is a challenging dataset designed to expose vulnerabilities in machine learning model performance. Created through a simple adversarial filtration technique that minimizes spurious cues, ImageNet-A presents a formidable challenge for existing models, surpassing the difficulty level of the conventional ImageNet \cite{deng2009imagenet} test set.
        Notably, a DenseNet-121 \cite{huang2017densely} model achieves a mere 2\% accuracy on ImageNet-A, reflecting a drastic 90\% drop in performance. The dataset comprises real-world, unmodified examples that consistently challenge diverse models, unveiling shared weaknesses in computer vision algorithms.

        \paragraph{ImageNet-R} \cite{hendrycks2021many} is a novel test set comprising 30,000 images that offers a distinctive challenge for evaluating the robustness of machine learning models. This dataset includes diverse renditions of ImageNet object classes, such as paintings and embroidery, introducing natural variations in textures and local image statistics not present in conventional ImageNet \cite{deng2009imagenet} images. By incorporating these naturally occurring renditions, ImageNet-R allows for a meaningful assessment of model performance in the face of realistic visual variations. The dataset serves as a valuable benchmark to gauge the effectiveness of previously proposed methods aimed at enhancing out-of-distribution robustness. Researchers can leverage ImageNet-R to rigorously test and compare various strategies for improving model performance on real-world, visually diverse renditions, offering a more comprehensive evaluation of robustness in the realm of image classification.

        \paragraph{ImageNet-C} \cite{hendrycks2019robustness} is a dataset designed to evaluate the robustness of machine learning models to various common visual corruptions. Comprising a collection of 75 widely encountered visual corruptions, this dataset applies these distortions to images from the ImageNet \cite{deng2009imagenet}. The introduction of ImageNet-C aims to establish a standardized benchmark for assessing the robustness of models to image corruptions, addressing concerns related to shifting evaluation criteria and cherry-picking results. By systematically subjecting images to a diverse set of corruptions, the dataset provides a comprehensive framework for benchmarking the performance of current deep learning systems. The findings from the evaluation underscore the considerable room for improvement in the robustness of models when confronted with the challenges presented by ImageNet-C.
    
    \subsection{Vision Transformer}

        Vaswani \textit{et al.} \cite{vaswani2017attention} first introduces the Transformer, a revolutionary architecture solely based on attention mechanisms, showcasing superior performance in machine translation tasks.
        Transitioning to computer vision, Vision Transformer (ViT) \cite{dosovitskiy2020image} challenges the convention of coupling attention with convolutional networks, proposing a direct sequence-based alternative based solely on attention mechanisms. ViT excels in image classification while demanding fewer computational resources.
        Swin Transformer \cite{liu2021swin} further refines transformer architecture for vision tasks, introducing a hierarchical design and a shifted window approach, yielding state-of-the-art results in image classification, object detection, and semantic segmentation.
        While ViT requires extensive pre-training, Steiner \textit{et al.} \cite{steiner2021train} proposed a novel approach to minimize training costs. They conducts an empirical study on Vision Transformers, highlighting their competitive performance with augmented regularization and increased compute, even when trained on smaller datasets.
        DeiT \cite{pmlr-v139-touvron21a} relies on knowledge distillation to reduce training costs. They demonstrates competitive results with a teacher-student strategy and introducing token-based distillation for effective knowledge transfer in attention-based models.
\end{document}